\RequirePackage[reqno]{amsmath}
\documentclass[letterpaper, 10 pt, journal, twoside]{ieeetran} 
\IEEEoverridecommandlockouts                       
\usepackage{svg}

\usepackage{svg}
\usepackage{color}
\usepackage{transparent}
\usepackage{graphicx}
\usepackage{microtype}  
\usepackage{import}
\usepackage{wrapfig}
\usepackage{algorithm}
\usepackage{diagbox}
\usepackage{subcaption}
\usepackage{multicol}
\usepackage{tablefootnote}
\usepackage{cases}
\usepackage{CJK}
\usepackage{indentfirst}
\usepackage{empheq}

\input{abkuerzung.sty}

\IEEEoverridecommandlockouts                              

\usepackage[utf8]{inputenc}
\usepackage{leftidx}
\usepackage{bm}
\usepackage{amssymb,amsmath,amsfonts,empheq}
\usepackage{graphicx}
\usepackage{siunitx}
\usepackage{multicol}
\usepackage{algorithm}
\usepackage{algpseudocode}
\usepackage{type1cm}
\usepackage{pifont}
\usepackage{lettrine}
\usepackage{color}
\usepackage{cite}
\usepackage{breqn}
\usepackage{tikz}
\usepackage{array}
\usepackage{import}
\usepackage{amsmath}
\usepackage{amssymb}
\usepackage{bm} 
\usepackage{booktabs}
\usepackage{optidef}
\usepackage{graphicx}
\usepackage{xcolor}
\usepackage{siunitx}
\usepackage{fixltx2e}
\usepackage{multirow}
\usepackage{cite}
\usepackage{import}
\usepackage[draft]{hyperref}
\usepackage{microtype}
\usepackage{tabularx}
\usepackage{booktabs}
\usepackage{comment}
 \usepackage{microtype}
 \usepackage{xifthen}

\usepackage{multirow}
\usepackage{scalerel}
\usepackage{fixltx2e}

\usepackage{xfp}
\usepackage{makecell}

\makeatletter
\renewcommand{\section}{\@startsection{section}{1}{\z@}{0.5ex plus 0.5ex minus 0.3ex}%
	{0.4ex plus 0.6ex minus 0ex}{\normalfont\normalsize\centering\scshape}}%
\renewcommand{\subsection}{\@startsection{subsection}{2}{\z@}{0.5ex plus 0.5ex minus 0.4ex}%
	{0.3ex plus 0.5ex minus 0ex}{\normalfont\normalsize\itshape}}%
\makeatother

\renewcommand{\baselinestretch}{0.905}
\global\long\def\dq#1{\underline{\boldsymbol{#1}}}%

\global\long\def\SE#1{\ensuremath{SE(#1)}}%

\newcommand{\distancesphere}{d^{S}}
\newcommand{\robotpoint}{\point^R}

\newcommand{\ctperstepms}[4]{%
  \fpeval{round(1000*#1/#3,4)} /\textcolor{gray}{\fpeval{round((1000*#1/#3)*sqrt((#2/#1)^2 + (#4/#3)^2),4)}}
}

\newcommand{\lf}[1]{{\textcolor{blue!70!red!60}{\textbf{LF:} #1}}}

\newcommand{\tofullversion}[1]{}  

\begin{document}

\title{GeoPF: Infusing Geometry into Potential Fields for Reactive Planning in Non-trivial Environments\vspace{-4pt}}

\author{Yuhe Gong$^{1}$, Riddhiman Laha$^{2}$, and Luis Figueredo$^{1,2,*}$\vspace{-11pt}
\thanks{$^{1}$The authors are with the School of Computer Science, University of Nottingham, Nottingham, United Kingdom.$^{2}$The authors are with Technical University of Munich (TUM), Munich Institute of Robotics and Machine Intelligence (MIRMI), Munich, Germany. *Corresponding author.
Email: \texttt{yuhe.gong@nottingham.ac.uk, riddhiman.laha@tum.de, figueredo@ieee.org}.   
}
}


\newcommand{\revised}[1]{\textcolor{black}{#1}}
\newcommand{\minorrev}[1]{#1}
\newcommand{\removed}[1]{}

\maketitle
\begin{abstract}

Reactive intelligence remains one of the cornerstones of versatile robotics operating in cluttered, dynamic, and human-centred environments. Among reactive approaches, potential fields (PF) continue to be widely adopted due to their simplicity and real-time applicability. However, existing PF methods typically oversimplify environmental representations by relying on isotropic, point- or sphere-based obstacle approximations. In human-centred settings, this simplification results in overly conservative paths, cumbersome tuning, and computational overhead---even breaking real-time requirements. In response, we propose the Geometric Potential Field (GeoPF), a reactive motion-planning framework that explicitly infuses geometric primitives---points, lines, planes, cubes, and cylinders---their structure and spatial relationship in modulating the real-time repulsive response. 
Extensive quantitative analyses consistently show GeoPF's higher success rates, reduced tuning complexity (a single parameter set across experiments), and substantially lower computational costs (up to 2 orders of magnitude) compared to traditional PF methods. Real-world experiments further validate GeoPF’s reliability, robustness, 
and practical ease of deployment, as well as its scalability to whole-body avoidance.  GeoPF provides a fresh perspective on reactive planning problems driving geometric-aware temporal motion generation, enabling flexible and low-latency motion planning suitable for modern robotic applications.

\end{abstract}
\IEEEpeerreviewmaketitle

\section{Introduction}

\begin{figure}[t]
    \centering
    \includegraphics[width=0.235\textwidth]{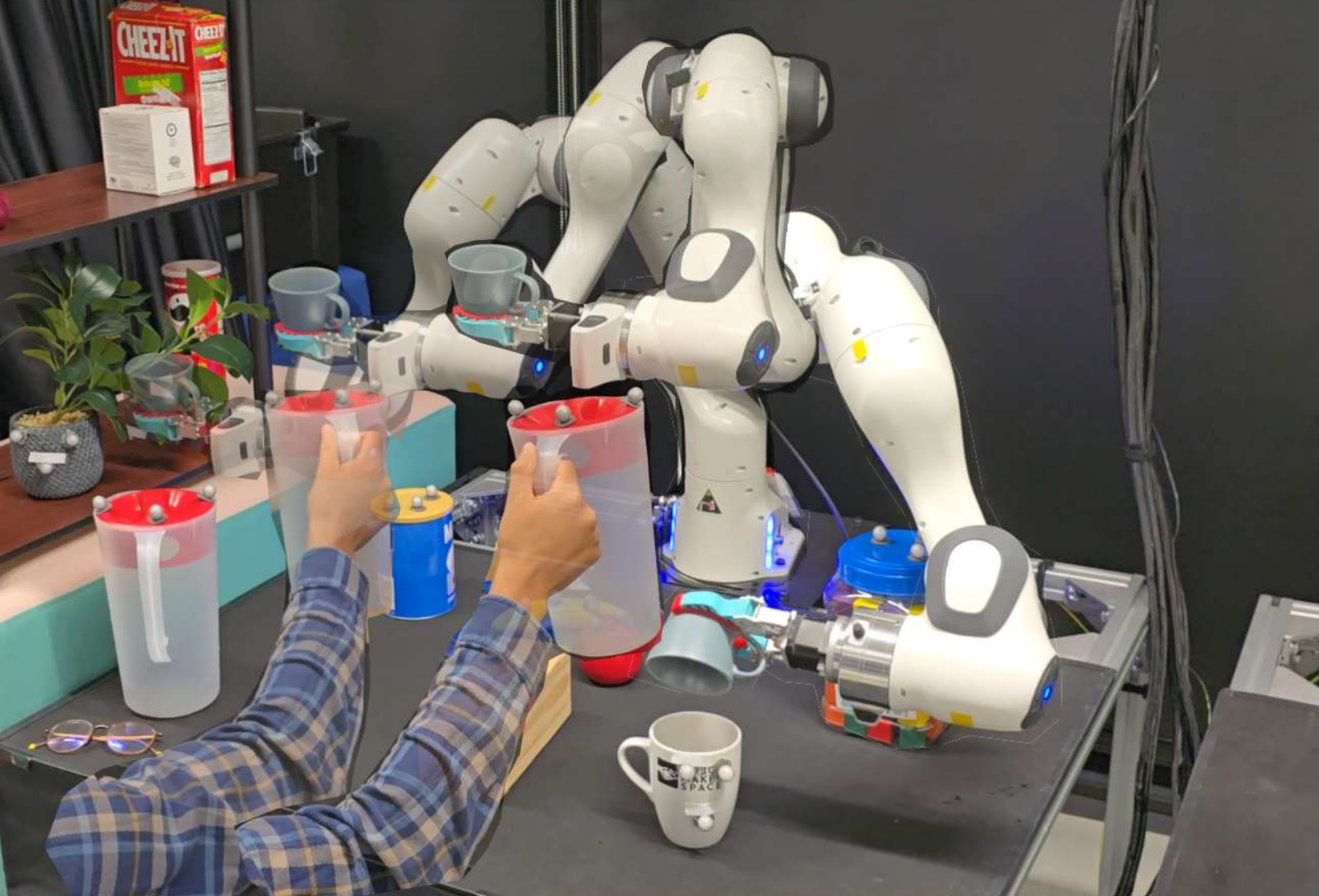}
    \includegraphics[width=0.235\textwidth]{figs/robot_pouring_water_back_compressed.pdf}
    \caption{Reactive robot behaviour in unstructured, cluttered, dynamic environments. 
    The robot transfers and pours the tea from the shelf to the cup (\textit{left}), 
    then returns the set while reacting to a human dynamically moving objects (\textit{right})---requiring accurate geometric representations for real-time motion adaptation.
    }
    \label{fig:teaser}
\end{figure}

Robotics has expanded beyond structured, well-defined factory floors into (i) unstructured, (ii) cluttered, and (iii) dynamic human-centred spaces, increasing the demand for reactive intelligent navigation, Fig.~\ref{fig:teaser}. 
Yet, despite significant advances in motion planning \cite{orthey2023sampling}, autonomous reactive behavior 
remains an open problem 
\cite{kappler2018real,laha2023predictive}.
The core challenge lies in 
\revised{addressing clutter while providing structured environment representations that enable} 
 %
real-time reactivity to dynamic, unforeseen changes~\cite{du2010robotic,marcucci2024fast}. 
This challenge is exemplified in Fig.~\ref{fig:teaser}. 
A robot must navigate a constrained environment to pour tea, set a 
table, and dynamically respond to a human moving objects in real-time. 
%
This 
requires a structured 
environment representation---\revised{one capable of seamlessly integrating geometric, spatial, and temporal information into real-time reactive planning.} 
This work addresses precisely this open challenge by introducing 
\revised{a novel paradigm that explicitly leverages shape-awareness, geometric structure, and spatial relationship into repulsive strategies beyond single-gradient avoidance---enabling}  
reactive intelligence, the centerpiece of truly versatile systems operating in human-centered non-trivial environments \cite{bryson2001modularity}.

Traditional monolithic Sense-Plan-Act (SPA) architectures divide the system into sequential sensing, planning, and action phases \cite{orthey2023sampling}. This strong modularization enables robots to function effectively in structured, pre-modeled environments \cite{DONALD1987295,correll2018analysis,kortenkamp2016robotic}. The key advantage 
lies in breaking down complex tasks into manageable subproblems, making planning more intuitive and computationally feasible \cite{kappler2018real}.  Most successful approaches including graph-based planners 
construct a free configuration space (C-space) representation and connect edges using local planners \cite{orthey2023sampling,lavalle2001randomized,kortenkamp2016robotic}. 
Instead, tree-based single-query planners, 
as Rapidly-exploring Random Trees (RRTs), iteratively expand toward the goal through
collision-free paths \cite{lavalle2001randomized}. Despite their effectiveness in structured settings, 
they  
lack real-time adaptability. Indeed, 
SPA methods excel in handling unstructured (i) and cluttered (ii) scenarios, but they fail to adapt efficiently to dynamic (iii) changes \cite{kappler2018real,laha2023predictive}. Their inherent sequential nature hinders real-time feedback, making them ill-suited for scenarios where reactive behavior and continuous adaptation to environmental changes are critical \cite{marcucci2023motion}.

%
%

In contrast, vector-field-based motion generation, or reactive planning,  
\revised{efficiently addresses the real-time responsiveness challenge (iii).} 
These methods 
use gradient-based techniques to guide motion generation \cite{Latombe1991,Warren1990,koren1991potential}. 
While early implementations suffered from local minima, 
these have been mitigated by 
strategies such as circular fields (CFs) \cite{singh1996real,haddadin2011dynamic,becker2023motion,laha2023predictive} and magnetic fields \cite{hussein2002motion,vallivaara2011magnetic,ataka2022magnetic}, \revised{which incorporated the sphere geometry to generate tangential repulsive forces to complement the standard repulsive gradient}. 

\revised{While mitigating traditional drawbacks (e.g., local minima), these methods  fundamentally rely on point or sphere-based obstacle representations.}  
\revised{This contrasts with modern simulators, engines and perception algorithms for collision detection (see e.g. MuJoCo \cite{todorov2012mujoco}, GJK  \cite{montaut2024gjk++}, and FCL \cite{pan2015hppfcl}) 
that compute (analytically or iteratively) contact-point distances and normals from richer geometric information.}  
\revised{Likewise, voxel- and Signed Distance Function (SDF)-based methods \cite{oleynikova2017voxblox,curobo_icra23,bien2024generating} 
(as well as neural SDFs variants \cite{staroverov2024dynamic}) 
compute geometry-aware closest-point smooth gradients for collision avoidance.  
}
\revised{Notwithstanding, during motion generation, all the aforementioned techniques reduce the obstacle interaction to a scalar distance with a repulsive to the closest point alone \cite{eckhoff2023towards} or local gradients \cite{bien2024generating} with respect to the closest point. That is, the rich information about the obstacle geometry, structure and spatial configuration is only used to compute the closest point and then neglected during motion generation.   
%
}


%


\revised{In other words, despite }
their computational efficiency compared to SPA, 
\revised{existing reactive planners remain fundamentally constrained to a single-point or single-gradient, towards the closest point, interaction---disregarding the richer geometry and structural information inherent to the object.}  
While intuitive, this particle-based approach \revised{provides poor information for the motion generation strategy. When considering point-clouds or point-based geometry, these methods lead to either large repulsive spheres yielding overly conservative plans, or dense sampling, which increases computational load}, making real-time execution impractical.  Conversely, sparse sampling causes force discontinuities, switching strategies, and short-sighted environment representations, ultimately leading to suboptimal or infeasible trajectories that are highly sensitive to tuning parameters.  

In summary, despite the promise of field-based planners in providing intuitive and reactive behaviours, their application 
remains constrained by three fundamental issues: (i) high computational costs associated with representing large, non-isotropic vector fields; (ii) limited performance due to the reliance on isotropic,
\revised{or closest-point-based distance fields and gradients;}
and (iii) the complex and impractical tuning requirements necessary to handle real-world scenarios effectively. 
These shortcomings are inherently tied to the environment representation upon which existing field-based methods are designed. In most cases, 
\revised{repulsive forces are build upon }
point clouds \revised{based distances} \cite{becker2023motion,laha2023predictive}, or through spheres or octomaps \cite{braquet2022vector,radhakrishnan2024state}, leading to challenges in handling non-spherical objects and non-isotropic vector fields. 
\revised{
%
While geometric-aware distance computation methods, e.g., \cite{todorov2012mujoco,montaut2024gjk++,pan2015hppfcl,oleynikova2017voxblox,curobo_icra23,bien2024generating} 
help reduce the computational load (i), the resulting   
%
reactive planners 
still ultimately reduce obstacles interactions to a single pointwise gradient, using the nearest-distance vector irrespective of whether the closest-contact lies on a face, an edge, or a corner.} 
\revised{As a result, planners must rely on local contact information only, which limits geometric expressiveness and leads to non-intuitive force profiles. This lack of structural awareness often requires }
non-intuitive and cumbersome tuning 
leading to poor or even infeasible trajectory generation.

\noindent\textbf{Contributions} 


\revised{The key novelty of this work lies in exploiting the rich structure of  geometric primitives to directly shape the reactive vector field.  
Rather than relying solely on the closest-point distance, 
GeoPF leverages the structural representation of each primitive. 
%
These features---promptly available in our method and other
geometric modeling pipelines---are most often overlooked during the planning stage. 
Similarly, in motion optimization frameworks, 
focus lies not on the shape itself but on how far the robot is from violating a closest-point contact-based threshold.
Thus, local point-based gradients dictate the collision logic instead of distinguishing between the robot approaching a face, corner, or a cylindrical wall. To the best of our knowledge, such information indeed has never been used during reactive planning and field generation.}



Our key contributions include: \revised{%
\textbf{(C1)} A novel geometric-aware planning paradigm, revising and generalizing primitives, leveraging their topological and structural properties to actively modulate the repulsive field  beyond simple distance-based computations; 
\textbf{(C2)} An analytical, closed-form vector-field planner explicitly tailored for common primitives (points, lines, planes, cubes, and cylinders), directly instantiating C1; 
\textbf{(C3)} Comprehensive quantitative analysis highlighting GeoPF's higher success rates and orders-of-magnitude computational improvements over state-of-the-art field-based planners, accompanied by real-world experiments --- including both whole-body avoidance and poor/coarse primitive shaping showcasing its scalability and robustness. 
}

\section{GeoPF: System Design and Overview}
\begin{figure}[t]
\centering
\includegraphics[width=1\linewidth]{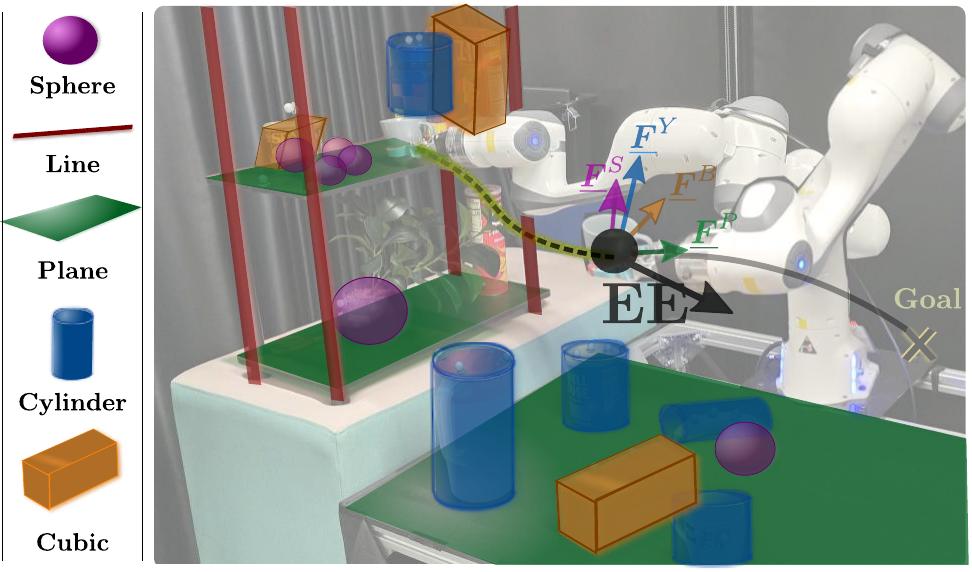}
\caption{
A snapshot of our GeoPF planner in the same setup as Fig.~\ref{fig:teaser}(\textit{left}). Each obstacle, captured as a sphere, line, plane, cube, or cylinder, generates a distance-based repulsive force. The coloured arrows illustrate the aggregated forces---per geometric type---steering the end-effector (EE) safely to the goal without collision.  
}
\label{fig::Overview}
\end{figure}
This work is about reactive planning in non-trivial human-centric environments. 
$ \dq{x} {\in} \SE{3} $. 
Let's consider an initial and a goal pose, respectively $\pose^0, \pose^g {\in} \SE{3}$. Both rotational and translational components are modelled through unit-dual quaternion parametrization, and while the orientation tracking is achieved through screw interpolation \cite{laha2022coordinate}, 
in this work, we are interested in the Cartesian trajectory, 
$\robotposition {\in} \realspace$.   


We list all additional notations used herein in Table~\ref{table::notation}. 
Furthermore, to improve the mathematical description, we will explore the following functions.  
\begin{align}
    \normalization(\point)  {=} \tfrac{\point}{||\point||},  \quad  
     \normaldirection(\point_1, \point_2)  {=} \normalization(\point^L_{1} {-}\point^L_{2}) = \tfrac{(\point^L_{1} {-} \point^L_{2})}{||(\point^L_{1} {-} \point^L_{2})||}.
    \label{eq:distance unitary}
\end{align}
where $\normalization(\point)$  and $\normaldirection(\point_1, \point_2)$  are the unit direction along $\point \in \realspace$ and from  $\point_2 $ pointing to $\point_1$. 


\begin{table}[t]
\centering
\caption{Notations used throughout the paper}\vspace{-5pt}
\label{table::notation}
\begin{tabular}{cc}
    \toprule
    Notation & Description \\ \midrule
    $<\cdot>$, $\times$  & Dot (inner) and Cross products \\
    \midrule
    ${Geo, S,L,P, B, Y}$ & Geometric Primitives (Geo),  Sphere (S),   \\
      &Line (L), Plane (P), Cubic (B), Cylinder (Y)  \\ 

    $\workspace$ & Robot workspace in $\realspace$ \\
    $\workspace^{obs}$ & Geometric Primitive Set  \\
    
    $\workspace^{feas}$, $\workspace_{\varnothing}$ & Feasible and Unfeasible Workspace  \\
    \midrule
    $\acceleration, \velocity, \point$ & Acceleration, velocity,  position in Cartesian Space \\
    $\point^R$ & Robot Position in Cartesian Space \\ 
    \midrule
    $\ndirection$, $\fdirection$ & Unit and Force Unit  direction  in Cartesian Space \\
    $\Force$ & Force \\
    $r^{Geo}$, $\point_{O}^{Geo}$ & Radius and Center of sphere, circle, cylinder \\
    
    \midrule
    \bottomrule
\end{tabular}
\end{table}


\noindent\textbf{Overview of the Design Strategy}

Given a convoluted obstacle geometric shape, simpler primitives can be fitted to achieve interactive performance. Herein, we make use of---spheres (balls, rounded objects), lines (wires, thin poles), planes (walls, floors), cubes (boxes, crates), and cylinders (pipes, cans)---as these geometric types effectively cover the majority of objects (we elucidate some of them) encountered in real-life scenarios. In the same vein, we designate the space occupied by each type as the spherical space $\workspace^{S}$, the line space $\workspace^{L}$, the plane space $\workspace^{P}$, the cubic space $\workspace^{B}$, and the cylinder space $\workspace^{Y}$. $
\workspace_{obs}$ can now be explicitly defined as $\workspace^{S} \cup \workspace^{L} \cup \workspace^{P} \cup \workspace^{B} \cup \workspace^{Y}$. 
Thus,
considering the full workspace of the robot $\workspace \subset \SE{3}$, 
the feasible space $\workspace^{feas}$ in a non-trivial environment is $\workspace^{feas} = \workspace \setminus (\workspace_{obs} \cup \workspace_{\varnothing})$, where   $\workspace_{\varnothing}$ designates the infeasible workspace, e.g., singular-poses, self-collision, joint limits, and workspace out-of-boundaries. 

Each geometric primitive is defined by a set of points that describe its spatial configuration in $\SE{3}$. Given the geometric shape and pose information, a geometric model can be constructed for each obstacle in the environment. One key feature to this aim is the normal vector, denoted as $\ndirection^{Geo}$, where $Geo \in \{S, L, P, B, Y\}$ represents the specific geometric type.

For each geometric primitive, exploring $\ndirection^{Geo}$, we are able to project the robot onto the obstacles. The projected point is the perpendicular foot $\point^{Geo}_{\bot R}$. Such a foot system indicates the spatial relationship between them, which in turn defines the avoidance strategy, as depicted in Figs.~\ref{fig::ptl::Field}-\ref{fig::ptY::Force}.  

 The core principle of potential fields is to model the robot and environment with attractive and repulsive potentials. The attractive force is along the direction from robot pointing to goal position $\fdirection^{attr} = \normaldirection(\point_g, \robotposition).$ The magnitude of the force is controlled by a tuning parameter $k^{attr}$. The attractive force exerted on the robot is given by $\Force^{attr} = k^{attr}\fdirection^{attr}.$ Differently, the repulsive force $\fdirection^{Geo}$ respects geometric considerations and it is directed along the shortest path from the obstacle to the robot. The magnitude of the repulsive force depends on the length of shortest distance $d^{Geo}$ with a tuning parameter $k^{Geo}$: 
\begin{align}
    \Force^{Geo} = \frac{k^{Geo}}{d^{Geo}}\fdirection^{Geo}.
\end{align}
Therefore, the force is directly proportional to the obstacle distance. The repulsive force should be only active when the robot is close to the obstacle. We set a threshold $\threshold^{Geo}$ for the activation. 
$\Force^{\varnothing}$ stems from additional workspace limit-based forces such as self-collision, joint limits, singularities, and workspace limits. The limit-based forces can be expressed through geometric primitives, $\workspace^{\varnothing} \subseteq \workspace^{Geo}$. An example is shown in Fig.~\ref{exp::fig::real}, where boundary limits restrict 
the feasible workspace.

\noindent\textbf{Problem Statement} The robot motion planning problem that we consider is hinged on the notion of potential functions. After the field construction we present earlier, the resultant net force acting on the robot can be expressed as
\begin{align}
    \Force^{res} = \Force^{attr} + \sum_{Geo} \Force^{Geo} + \sum_{\varnothing}\Force^{\varnothing},
\end{align}
The robot's acceleration at each timestep can be computed as $\acceleration_i = \Force^{res}_i / \mass$, where $\mass$ denotes the mass of the robot, and $i \in \{1, 2, 3, ...\}$ is the timestep index. The velocity will be updated through $\velocity_{i+1} = \velocity_i + \timeinterval\acceleration_i$, $\timeinterval$ being the step size. Thus, the Cartesian robot position updates as
\begin{align}
    \point^R_{i+1} = \point^R_{i} + \frac{1}{2} \timeinterval^2\acceleration_i + \timeinterval \velocity_i.
\end{align}
Lastly, the orientation of the robot is determined by the task requirements, such as the grasping direction. \textit{Thus, we seek a  feasible trajectory $\feasiblepath$ starting from $\pose^0$ and ending in  $\pose^g$ in the free task space} $\workspace^{feas}$ \textit{brought about by $\Force^{res}$ and taking into account the workspace (environmental) constraints as well as geometrical considerations}.

\section{From Geometry to Force Generation}\label{sec:geometry_to_force}

One of the key contributions of this work is a geometric-based environment representation that explicitly encodes the spatial properties of objects and uses this geometry to drive temporal motion generation. Rather than relying solely on point clouds or large sets of spheres for obstacle modeling, \revised{or the closest-point contact-based to the obstacle,} we cast primitives—including spheres, lines, planes, cubes, and cylinders—into repulsive forces within a vector field.


Formally, let $\{G_{i}\}$ denote the set of geometric primitives representing the obstacles. Each primitive $G_{i}$ is associated with its own distance function $d_{i}(\cdot)$ and corresponding repulsive force $\Force_{i}$. The aggregate motion is determined by summing these repulsive forces with an attractive force pulling the robot toward the goal $\point^{g}$. Specifically,
\begin{equation}
\Force^{\mathrm{res}} = 
    \underbrace{k^{attr}\fdirection^{attr}}_{\text{goal attraction}}
+
\sum_{i}
\negthickspace   \negthickspace  \negthickspace \negthickspace 
    \underbrace{
    \left(\frac{k_i}{\,d_i(\cdot)\,} \fdirection_i\right)}_{\text{repulsion from obstacle }i}
\negthickspace  \negthickspace \negthickspace
+ \ 
   \sum_{j} \Force^{\varnothing}_j,
\label{eq:f_res}
\end{equation}
where $k_{\mathrm{attr}}$ and $k_i$ are scalar gains.
%
The following subsections detail how each geometric primitive defines its distance function $d_{i}(\cdot)$ and force direction~$\mathbf{f}_i$.

\subsection{\textbf{Point and Sphere-based Potential Field}}

A sphere is defined by its centroid  $\spherecenter {\in} \realspace$ and radius $\sphereradius$, while a point is given by $\sphereradius {=} 0$. It forms the simplest geometric primitive we define for collision avoidance. First, we compute the \textit{shortest distance} ($\distancesphere$) between the point robot and the sphere’s surface, i.e., $d^S {=} || \robotpoint - \spherecenter|| - \sphereradius$.  
Second, we are interested in the direction of the relative position vector, i.e., 
the radial vector outward from the sphere center,  $\fdirection^{S} {=} \normaldirection(\point^R, \spherecenter) $. 
%
 The repulsive force is therefore computed as 
\begin{equation}
\Force^{S} = \tfrac{k^{S}}{d^{S}}  \fdirection^{S} , 
\quad 
\text{where } 
\fdirection^{S} {=} \tfrac{ \point^R {-} \spherecenter  }{ || \point^R {-} \spherecenter || },  
\ \    
k^{S} {>} 0. 
\end{equation}
The sphere-based repulsive force can be enhanced by circular forces or magnetic-based \cite{ataka2022magnetic,laha2023predictive}, yet we explicitly decided to maintain a straightforward approach, avoiding unnecessary complexity of the spherization---that often rely on tuning for efficiency, as shown in Section \ref{sec:QuantitativeAnalysis}. For more details  on alternative forces, please refer to \cite{laha2023predictive,becker2023motion}. 





\subsection{\textbf{Segment Line-based Potential Field}}
\label{sec::linefield}
In the remainder of this section, we present additional geometric-primitive-based potential field generation. Starting from a line segment, defined 
by two vertices 
$\point^L_{1}, \point^L_{2} {\in} \realspace$.

\begin{figure}[t]
\centering
\begin{subfigure}{0.49\linewidth}
    \centering
    \includegraphics[width=1\linewidth]{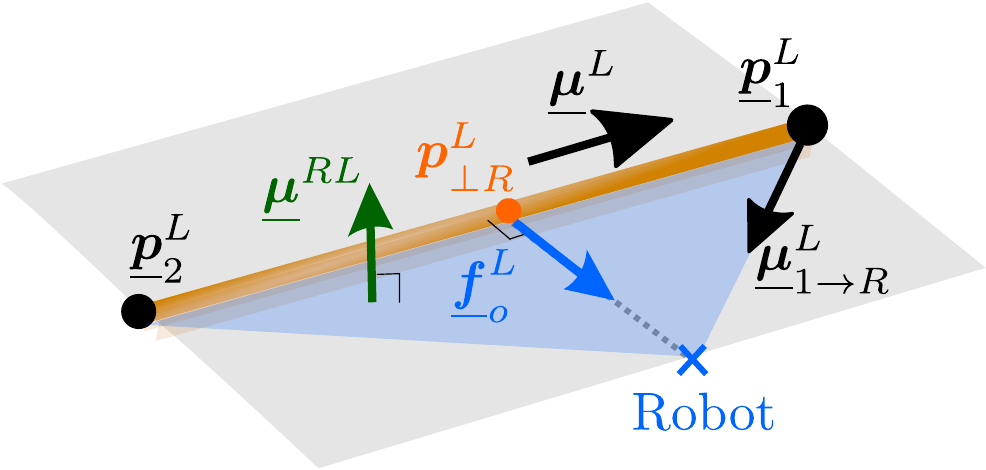}
    \caption{Orthogonal Line Field}
    \label{fig::ptl::Field1}
\end{subfigure}
\begin{subfigure}{0.49\linewidth}
    \centering
    \includegraphics[width=1\linewidth]{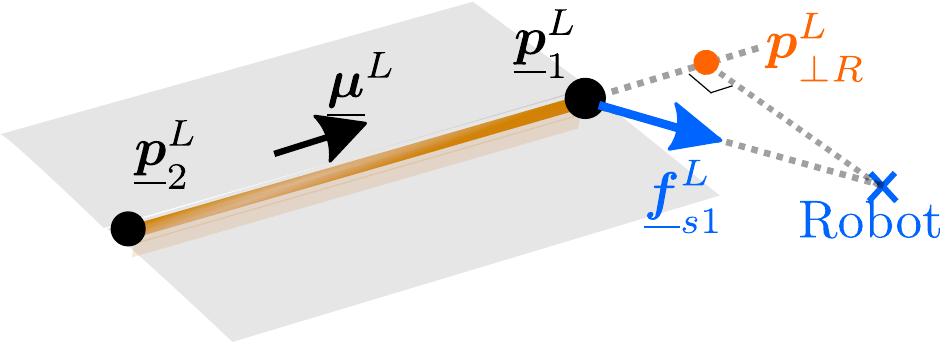}
    \caption{Side Line Field}
    \label{fig::ptl::Field2}
\end{subfigure}
\caption{%
Line-based potential field. (a) When the perpendicular foot is inside the line segment. The force is orthogonal. (b) When the perpendicular foot is outside the segment, the force is side force. }
%
\label{fig::ptl::Field}
\end{figure}

The repulsive force surrounding the line segment is a function of the shortest path from the line to the robot. The shortest distance, and thereafter, the repulsive force direction, between the robot ($\robotpoint$) and the finite segment may arise from one of two cases, as shown in Fig.~\ref{fig::ptl::Field}.  
\begin{enumerate}
    \item The robot projected onto the line segment, as shown in Fig~\ref{fig::ptl::Field1}, indicating that the perpendicular foot lies between the vertices. In the other words, the robot is orthogonal to the line segment. 
    \item Otherwise, the robot lies outside the orthogonal region, as shown in Fig~\ref{fig::ptl::Field2}. In the other words, the robot is on the side of the line segment.
\end{enumerate}
In the first case, the resulting orthogonal-based force direction is defined as $\fdirection^L_o$, whereas the side forces are given by $\fdirection^L_{s1}$ and $\fdirection^L_{s2}$, as shown in Fig.~\ref{fig::ptl::Field}.

To evaluate both conditions,  
we first compute the perpendicular foot that the robot projects onto the line.  
To compute it, we first 
obtain the robot-line plane based on the line within the plane, e.g.,  
$\ndirection^L_{1\to R} {=} \normaldirection(\point^R,\point^L_1)$, 
and the plane's normal direction, given by 
$\ndirection^{RL} {=} 
(\ndirection^L_{1\to R} {\times} \ndirection^L)$, where 
$\ndirection^L {=} \ndirection^L_{2\to1} {=} \normaldirection(\point^L_{1}, \point^L_{2})$ is the unitary vector from $\point^L_{2}$ towards $ \point^L_{1}$.  Next, we compute the orthogonal direction perpendicular to both the line and the plane normal, i.e., 
$\ndirection^L_o = 
(\ndirection^L {\times} \ndirection^L_R)$. 
Finally, the orthogonal distance from the robot to the line is given by    
$d^L_{o} = <\point^R {-} \point^L_1, \ndirection^L_o>$, which is the length of the projection of the robot to    $\point^L_1$ over  $\ndirection^L_o$.  
The perpendicular foot, in this case, also the projection point, can be calculated as $\point_{\bot R}^L = ( \point^R {-} d^L_{o} ) \ndirection^L_o$. 

When the perpendicular foot $(\point_{\bot R}^L)$ lies  between the vertices of the line segment, i.e., $\point_{\bot R}^L {\in} \workspace^L$,  
the repulsive force is the orthogonal force. 
In this case, the distance from the perpendicular foot to the edge is always smaller than the line distance, i.e., $d^L_1 < l^L \cap d^L_2 > l^L$. Note that the distance can be calculated using the Euclidean norm, given the known position of perpendicular foot and the vertices. 
Conversely, when the perpendicular foot is outside the line segment $\point_{\bot R}^L\not\in\workspace^L$, the repulsive force is the side force, directed from the closest vertices towards the robot:
\begin{subequations}
\begin{empheq}[left={(\fdirection^L, d^L) = \empheqlbrace}]{align}
& (\fdirection^L_{s1},d^{L}_{1}) & \text{$d^L_1 > l^L \cap d^L_2 > l^L$}, \label{lineforceA}\\
& (\fdirection^L_{o}, d^L_o) &  \text{$d^L_1 < l^L \cap d^L_2 < l^L$},  \label{lineforceB}\\
& (\fdirection^L_{s1},d^L_1) & \text{$d^L_1 < l^L \cap d^L_2 < l^L$}, \label{lineforceC} 
\end{empheq}
\begin{align}
    \Force^{L} = \tfrac{k^{L}}{d^{L}} \fdirection^{L}.   
    \label{lineforceD}
\end{align}
\end{subequations}


\subsection{\textbf{Plane-based Potential Field}}
\label{sec::planefield} 
Planes are among the most common geometric obstacles in human-centred environments---representing tables, walls, and other flat surfaces. Traditional methods approximate them using multiple discrete spheres, as shown in Fig.~\ref{fig:exp:plane scenario1}. Although conceptually simple, this results in inefficient and imprecise representations, requiring extensive tuning to prevent the field from either penetrating the spheres or generating unnecessarily large repulsive forces. In contrast, explicitly considering a plane with four vertices allows for closed-form distance and force computations, reducing both computational overhead and tuning complexity.  
Indeed, let's consider a plane defined by four (counter-)clockwise ordered vertices $\point^{P}_{1},\point^{P}_2,\point^{P}_3,\point^{P}_4 {\in} \realspace$. 
The plane's outward normal can be computed as 
\begin{align}
    \ndirection^{{P}} = \normalization(\ndirection_{2\to1}^P \times \ndirection_{2\to3}^P),
    \label{eq::planedirection}
\end{align}
where 
$\ndirection_{2\to1}^P=\normaldirection(\point^P_{1}, \point^P_{2})$ and $\ndirection_{2\to3}^P=\normaldirection(\point^L_{3}, \point^L_{2})$ depict segments of line within the plane.  

%

Similar to the segment-line case, 
the repulsive force, thus the avoidance strategy, is determined by the robot's projection onto the perpendicular foot ($\point_{\bot R}^P$) on the obstacle plane.  

The perpendicular foot on the plane is computed by  
$\point^P_{\bot R} = \point^R {-} d^P_o\ndirection^P$  
where 
$d^P_o = {<{\ndirection^P, \point^R {-} \point^P_1}>}$  
is the spatial distance between the plane and the robot projected on the orthogonal outward normal from \eqref{eq::planedirection}.
%

%
\begin{figure}[t]
\centering
\begin{subfigure}{0.49\linewidth}
    \centering
    \includegraphics[width=1\linewidth]{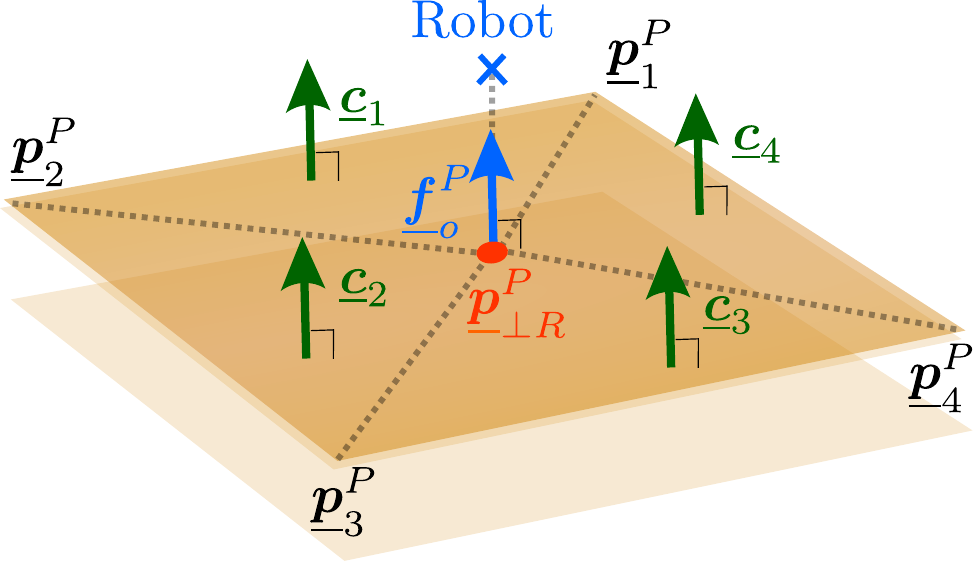}
    \caption{Orthogonal Plane Field}
    \label{fig::ptp::Field1}
\end{subfigure}
\begin{subfigure}{0.49\linewidth}
    \centering
    \includegraphics[width=1\linewidth]{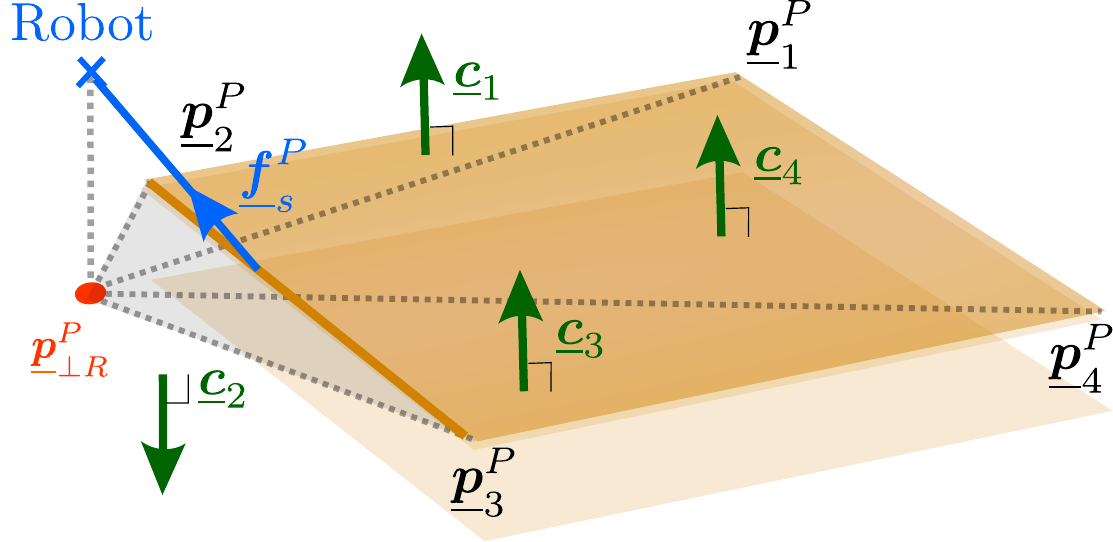}
    \caption{Side Plane Field}
    \label{fig::ptp::Field2}
\end{subfigure}
\caption{Plane-based potential field defined by the perpendicular foot $\point^P_{\bot R}$. (a) If it lies within the plane boundary, the condition vectors (green arrows) remain aligned, see \eqref{eq::pointinsideplane}, then the repulsive force direction  $\fdirection^P$ aligns with the plane’s normal. (b) Otherwise, the force is derived from the nearest edge.}
%
\label{fig::ptp::Field}
\end{figure}


Whether the perpendicular foot lies within a rectangle's plane can be evaluated by  
angles between the lines connecting the foot point and the rectangle's vertices. Let the direction from the perpendicular foot to the $i^{th}$ vertex be denoted as $\ndirection^P_{r i}$, where $i\in\{1,2,3,4\}$. If the perpendicular foot is located within the plane, the angles between two adjacent foot-vertex lines should be less than 180°. 
In contrast, if any of the angles exceed 180°, the point lies outside the plane.

The angle condition can be determined through the cross-product  
$< \negthinspace \ndirection^P_{r i}, \ndirection^P_{r( i+1)} \negmedspace >$. 
Therefore, we have the normal direction of the plane given by  
\begin{align}
   \boldsymbol{\underline{c}}_1 := 
        \normalization(\ndirection^P_{r 1} \times \ndirection^P_{r 2}),  
    \quad 
    \boldsymbol{\underline{c}}_2 := 
        \normalization(\ndirection^P_{r 2}\times\ndirection^P_{r 3}), \\
   \boldsymbol{\underline{c}}_3  := 
        \normalization(\ndirection^P_{r 3}\times \ndirection^P_{r 4}), 
    \quad 
   \boldsymbol{\underline{c}}_4 := \normalization(\ndirection^P_{r 4}\times \ndirection^P_{r 1}).
   \label{eq::pointinsideplane}
\end{align}
The perpendicular foot lies inside the plane if and only if all  
indicators $\boldsymbol{\underline{c}}_i$ point in the same direction, i.e.,  
$\boldsymbol{\underline{c}}_1  {=} 
\boldsymbol{\underline{c}}_2   {=}
\boldsymbol{\underline{c}}_3   {=}
\boldsymbol{\underline{c}}_4$, 
Fig.~\ref{fig::ptp::Field1}. 
In this case, the perpendicular foot lies inside the plane boundaries $\point^P_{\bot R} \in \workspace^P$, and the repulsive force direction is orthogonal to the plane and aligned to \eqref{eq::planedirection}, i.e.,  $\fdirection_o^P = \ndirection^P = \boldsymbol{\underline{c}}_i$.


When any of $\boldsymbol{\underline{c}}_i$ points in the opposite direction (Fig.~\ref{fig::ptp::Field2}), the point is outside the plane, $\point^P_{\bot R} \not \in \workspace^P$, which guarantees the convexity of the plane. The resulting side force $\fdirection_s^P$ and corresponding distance $d_s^P$ is then redirected from the nearest plane edge or vertex using line-based logic as in Sec.~\ref{sec::linefield}.

The resulting plane repulsion is computed as 
\begin{equation}  
\Force^P  = \tfrac{k^P}{\,d^P} \fdirection^P, 
\label{eq:plane:force}
\end{equation} 
where 
$(\fdirection^P, d^P)$ is given by either $(\fdirection_o^P,d_o^P)$ or $(\fdirection_s^P,d_s^P)$ according to the case described above.

\noindent\textbf{Plane-based Correction Potential Field Heuristic}

\begin{figure}[t]
\centering
\begin{subfigure}{0.49\linewidth}
    \centering
    \includegraphics[width=1\linewidth]{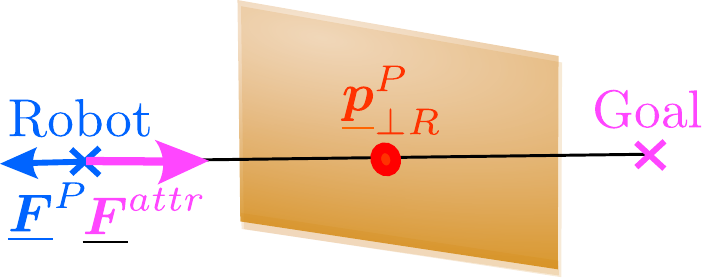}
    \caption{Fail to avoid the plane}
    \label{fig::ptp::Correction1}
\end{subfigure}
\begin{subfigure}{0.49\linewidth}
    \centering
    \includegraphics[width=1\linewidth]{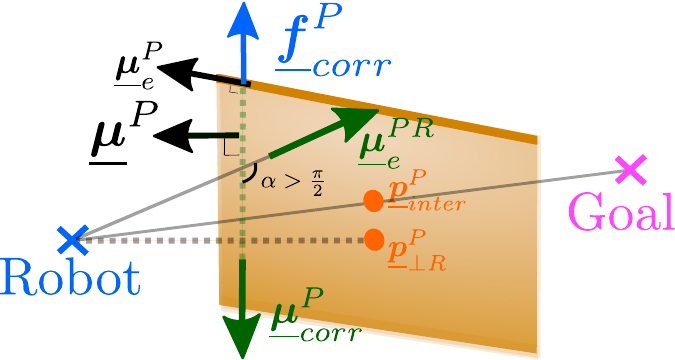}
    \caption{Plane Field Correction}
    \label{fig::ptp::Correction2}
\end{subfigure}
\caption{Plane force correction. (a) 
Without correction, a large plane may trap the robot by canceling the repulsive and attractive forces.
(b) 
When the line connecting the robot and goal intersects with the plane ($\point_{inter}^P$), the force is redirected to the parallel direction
$\ndirection^P_{corr}$. If the parallel direction points aways of the nearest edge, the force is redirected according to $\alpha$ with  $\fdirection^P_{corr}$ being the corrected  direction.}  
\label{fig::ptp::Correction}\vspace{-10pt}
\end{figure}

While successful in avoiding collision with plane-based geometries,
when a large plane is positioned between the robot and the goal, 
the attractive force counteracts the repulsive force, causing a trap condition (Fig.~\ref{fig::ptp::Correction1}). 

 If the line from robot to goal intersects the plane interior, we project a repulsive force parallel to the plane’s surface (toward the nearest plane edge) rather than purely along the normal. This prevents the attractor from canceling out the plane’s repulsion, allowing the robot to “slide” around or above the plane.

Given the unit vector along the robot-goal line, $\ndirection^{RG} = \normalization(\point^R-\point^G)$, 
the intersection point in the unbounded plane is given by 
$\point^P_{inter} {=} \point^R {-} d^P_{inter}\ndirection^{RG}$, 
where $d^P_{inter} {:=}  \frac{d^P_o}{<\ndirection^{RG}, \ndirection^P_o>}$ is the distance from the robot to the intersection point.
If the intersection point lies within the plane, we project a force parallel to the plane toward the nearest plane edge.

By defining the nearest edge direction $\ndirection^P_{e}$, 
the parallel direction is given by $\ndirection^P_{corr}=\ndirection^P\times \ndirection^P_{e}$. 
To avoid crossing 
toward the farthest edge, see Fig.~\ref{fig::ptp::Correction2}, 
we evaluate the direction from the robot to the nearest edge{\footnote{\revised{In rare case of exact symmetry, similarly to \cite{laha2023predictive}, a small random additive noise is applied toward one of
the primitive’s vertices or edges.}}}, 
denoted as $\ndirection^{PR}_{e}$, which corresponds to the opposite direction of the line-to-robot vector, Section~\ref{sec::linefield}. 
By computing  $s = \text{sign}(<\ndirection^P_{corr}, \ndirection^{PR}_{e}>)$, we can determine whether the parallel direction toward the nearest edge or its opposite. When $s=1$, the direction is correctly toward the nearest edge ($\alpha < \frac{\pi}{2}$), whereas  $s=-1$ indicates the farthest edge ($\alpha > \frac{\pi}{2}$), see Fig.~\ref{fig::ptp::Correction}. 
Accordingly, the corrected force is 
\begin{equation}
    \fdirection^P_{corr} = s\ndirection^P_{corr}, 
    \quad 
    \text{with} \ \ 
    \ndirection^P_{corr}{=}\ndirection^P {\times} \ndirection^P_{e},  
    \label{eq:plane:correction force}
\end{equation}
with the magnitude of the force according to the distance between the robot and the plane - the orthogonal force and the side force corresponding distance.

\tofullversion{
\lf{I like the description (but perhaps we leave it for the next journal. We are running out of space). Let's save a copy with all these expressions, but for the submitted version we cut it due to space:}
Summarized, the plane force can be expressed as
\begin{empheq}[left={(\fdirection^P, d^P) = \empheqlbrace}]{align}
& (\fdirection^P_{o},d^{P}_{o}) & \text{$\point^P_{\bot R},\in \workspace^T, \point^P_{inter}\notin \workspace^P$}, \tag{5a}\label{PlaneforceA}\\
& (\fdirection^P_{s},d^{P}_{s}) & \text{$\point^P_{\bot R},\notin \workspace^T, \point^P_{inter}\notin \workspace^P$},  \tag{5b}\label{PlaneforceB}\\
& (\fdirection^P_{corr},d^{P}_{o}) & \text{$\point^P_{\bot R}\in \workspace^P, \point^P_{inter}\in \workspace^P$}, \tag{5c}\label{PlaneforceC} \\
& (\fdirection^P_{corr},d^{P}_{s}) & \text{$\point^P_{\bot R}\notin \workspace^P, \point^P_{inter}\in \workspace^P$}, \tag{5d}\label{PlaneforceD} 
\end{empheq}
\begin{align}
    \Force^{P} = \frac{k^{P}}{d^{P}} \fdirection^{P}. \tag{5e}
    \label{PlaneforceE}
\end{align}
}


\textcolor{blue}{GeoPF has} orders-of-magnitude improvements in computational speed over
state-of-the-art reactive field-based planners while maintaining
higher success rates.


\subsection{\textbf{Geometric Cube-Based Potential Field}}
\begin{figure}[t]
\centering
\begin{subfigure}{0.48\linewidth}
    \centering
    \includegraphics[width=1\linewidth]{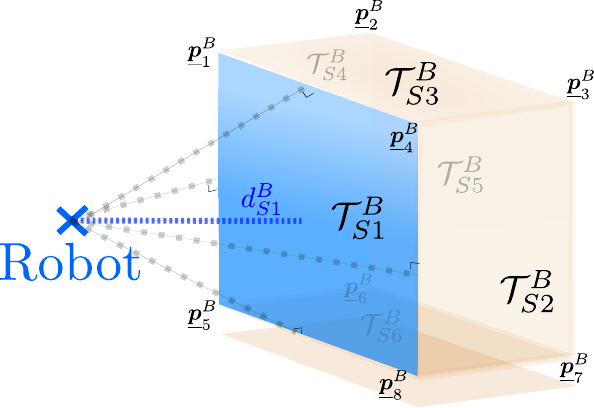}
    \caption{Cubic Surface Field on $\workspace^B_{S1}$}
    \label{fig::ptb1}
\end{subfigure}
\begin{subfigure}{0.48\linewidth}
    \centering
    \includegraphics[width=1\linewidth]{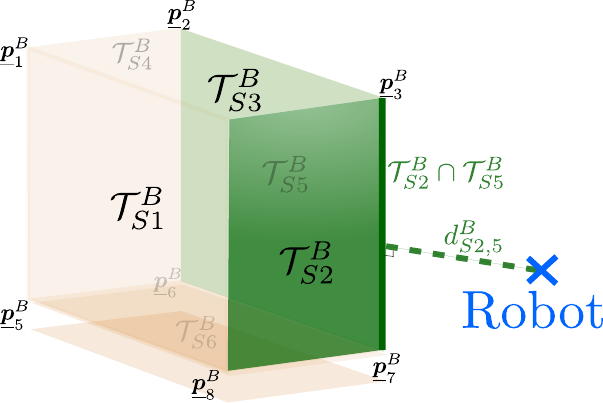}
    \caption{Cubic Edge Field on $\workspace^B_{S2}\cap \workspace^B_{S3}$}
    \label{fig::ptb2}
\end{subfigure}
\caption{Cubic force relies on the closest surface. (a) When robot closest to surface, the surface plane provides the force (Sec.~\ref{sec::planefield}. (b) When robot closest to an edge, either surface can be chosen.}
\label{fig::ptb::Force}
\end{figure}

Extending from previous primitives, 
many everyday objects, from boxes to shelves and cabinets, can be approximated as cubes or rectangular prisms. To model a cube, we use eight (counter-)clockwise ordered vertices $\point^{B}_{i} {\in} \realspace$, \ i{=}\{1,...,8\}.
The vertices  $\point^{B}_{1,2,3,4}$ represent one surface, whereas $\point^{B}_{5,6,7,8}$ representing the opposite surface. Moreover, $\point^{B}_{1}$ is aligned with $\point^{B}_{5}$, as illustrated in Fig.~\ref{fig::ptb::Force}. Therefore, the cube can be decoupled into six different surfaces $\workspace^B_{S1}$ to $\workspace^B_{S6}$ and their intersection edge $\workspace^B_{Si} \cap \workspace^B_{Sj}$, with $i\neq j \in \{1,2,3,4,5,6\}$. 

In practice, each of the six faces is treated like an individual plane (as in Section~\ref{sec::planefield}), so the force generation relies on choosing whichever face is nearest, as shown in Fig.~\ref{fig::ptb1}.
When the robot is closest to a cube edge $\workspace^B_{Si} {\cap} \workspace^B_{Sj}$, then the forces generated by $\workspace^B_{Si,j}$ are equivalent and the distance and direction of repulsion are handled via plane or line logic, whichever is more appropriate according to Section~\ref{sec::linefield}. In summary, 
\begin{align}
    \Force^{B} = \frac{k^{B}}{d^{B}} \fdirection^{B}, (\fdirection^B, d^B) =  (\fdirection^B_{Si},d^{B}_{Si}), i=\arg\min^6_{i=1} d^B_{Si}.
    \label{Cubicforce}
\end{align}


\subsection{\textbf{Cylinder-Based Potential Field}}
\label{sec:compute:cylinder}
\begin{figure}[t]
\centering
\begin{subfigure}{0.58\linewidth}
    \centering
    \includegraphics[width=.95\linewidth]{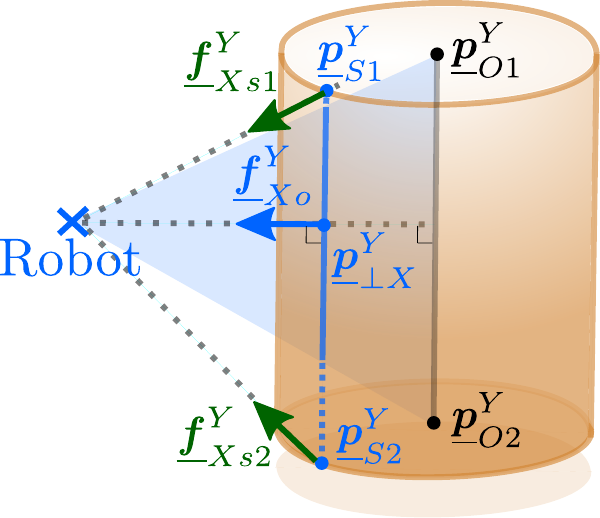}
    \caption{Surface Line Field}
    \label{fig::ptY1}
\end{subfigure}
\begin{subfigure}{0.38\linewidth}
    \centering
    \includegraphics[width=.95\linewidth]{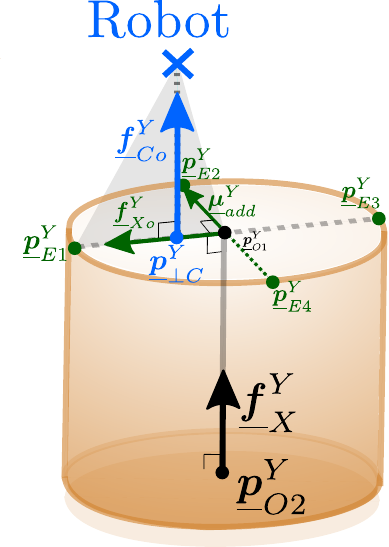}
    \caption{Circular Plane Field}
    \label{fig::ptY2}
\end{subfigure}
\caption{Cylinder Force has two different situations. (a) When the robot is close to the curved surface, the force can be regarded as a line force. (b) When the robot is close to the circular plane, the force can be regarded as the plane force. By checking the circular edge points, we can build the plane field on the circular plane.}  
\label{fig::ptY::Force}
\end{figure}

Cylinders commonly arise from objects such as cups, bottles, cans, pipes, or cylindrical machinery components. We define a cylinder by its central axis (stemming from two vertices $\point^Y_{O1},\point^Y_{O2} {\in} \realspace$) and the radius  $r^Y$.
%
We explore the curved surface surrounding the cylinder, and two circular-cross section planes 
centred at the vertices. 
The normal direction of the curved surface is the radial direction from the central axis 
extending outward, Fig~\ref{fig::ptY::Force}. The normal 
of the circular surfaces are perpendicular to the plane and point outside the cylinder, 
then we directly apply the force along the axis direction $\fdirection^Y_X$.

When the robot is “beside” the curved surface---as in Fig.~\ref{fig::ptb1}---the force can be computed akin to a line-based approach, whereas if it is above or below, we treat the top or bottom circle similar to the plane-like surface. 

Assuming no penetration, the former occurs when the distance ($d^Y_{\bot X}$) from the robot to its perpendicular foot within the cylinder-axis is larger than the radius, whereas $d^Y_{\bot X} {<} r^Y$ indicates the robot is right above or below the cylinder (Fig.~\ref{fig::ptY2}).

To compute the resulting forces, we compute a robot-axis plane and the orthogonal vector to the cylinder axis ($\ndirection^Y_{\bot X}$)---as in Section~\ref{sec::linefield}. Note $d^Y_{\bot X}$ is computed along $\ndirection^Y_{\bot X}$.


\noindent\textbf{Surface-Line Potential Field}

When the robot is “beside” the curved surface ($d^Y_{\bot X} > r^Y$), it is 
closest to a line segment on the surface of the cylinder. To this aim, we shift the cylinder axis to the closest point of the surface by translating the vertices along the direction $\ndirection^Y_{\bot X}$, i.e.,   
$\point^Y_{Si} {=} \point^Y_{Oi} + d^Y \ndirection^Y_{\bot X}$, 
$ \ i{=}\{1,2\}$. 
 
The resulting force is defined by the robot to the segment line $\overline{\point^Y_{S1}\point^Y_{S2}}$ as in Section~\ref{sec::linefield}. Thus, we have the surface-line orthogonal force $\fdirection^Y_{Xo}$, side force $\fdirection^Y_{Xs}$ and the surface-line perpendicular foot $\point^Y_{\bot X}$.

When the perpendicular foot of surface line $\point^Y_{\bot X}$ is in between $\point^Y_{S1}$ and $\point^Y_{S2}$, the cylinder force is the orthogonal force $\fdirection^Y = \fdirection^Y_{Xo}$, which is the same as the axis orthogonal vector $\fdirection^Y_{Xo} = \ndirection^Y_{\bot X}$. When the surface line foot $\point^Y_{\bot X}$ is outside, the cylinder force is the side force $\fdirection^Y = \fdirection^Y_{Xsi}$, with $i = \{1,2\}$. The distance is computed from the robot to surface's line.

%

\noindent\textbf{Circular-Plane Potential Field} 

When robot-axis distance is smaller than the radius ($d^Y_{\bot X} < r^Y$), the robot remains either on the top or bottom of the cylinder (Fig.~\ref{fig::ptY2}). When the circular plane perpendicular foot is inside the plane: $\point^Y_{\bot C}\in\workspace^{Y}_C: ||\point^Y_{\bot C}-\point^Y_{O1}|| < r^Y$, we apply the circular plane force. Otherwise, it is the surface line force. The circular plane normal is the unit vector along cylinder axis $\fdirection^Y_{X} = \normaldirection(\point^Y_{O1}, \point^Y_{O2})$. Note that we have surface points $\point^Y_{S1,2}$. Suppose the robot is closer to the upper surface with the surface point $\point^Y_{S1}$. Then it is easy to find another point on the circular plane by taking the cross product $\ndirection^Y_{add}=<\ndirection^Y_{\bot X}, \fdirection^Y_{Xo}>$. Then, the four points on the edge of the circular plane are $\point^Y_{E1,3} = \point^Y_{O1} \pm d^Y \ndirection^Y_{add}$, and $\point^Y_{E2,4} = \point^Y_{O1} \pm d^Y \ndirection^Y_{\bot X}$. 
%
With the three of them, we can shape a plane and calculate the plane force according to Sec.~\ref{sec::planefield}. However, a key distinction is that the angle-based condition for verifying the intersection point $\point^Y_{C_{\times}}$ is no longer applicable, as the original plane is circular. Instead, the intersection point can be checked by the distance. If the distance from the intersection point to the circle center exceeds the radius $\point^Y_{C_{\times}}\notin\workspace^{Y}_C: ||\point^Y_{C_{\times}}-\point^Y_{O1}|| > r^Y$, the intersection point lies outside the circular plane. Here, the distance is same as the plane-based distance for either correction or non-correction force.
\begin{figure}[t]
\centering
\begin{subfigure}{1\linewidth}
\label{fig:exp:plane scenario1}
    \includegraphics[width=\linewidth]{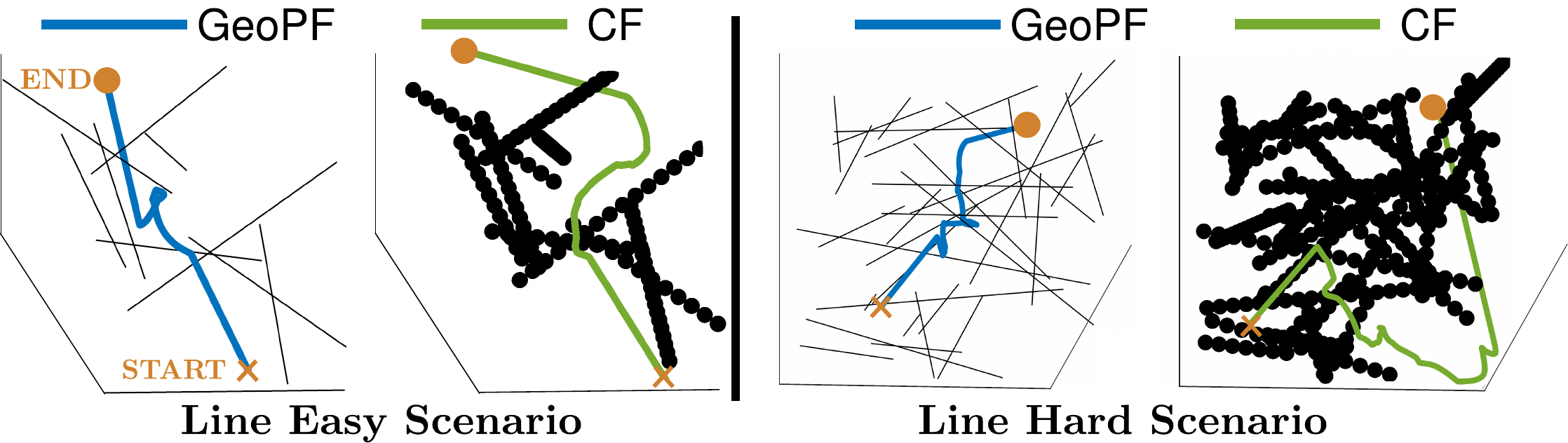}
    \caption{Comparison of GeoPF and CF Spherization in Easy (Left) and Hard (Right) Scenarios. 
    Black lines and circles represent line obstacles and their spherization. 
    In clutter, spherization forces trajectories around all obstacles, while GeoPF navigates 
    narrow spaces between obstacles and successfully reaches the goal.}  
    \end{subfigure}
    \begin{subfigure}{0.48\linewidth}
    \centering
    \includegraphics[width=0.8\linewidth]{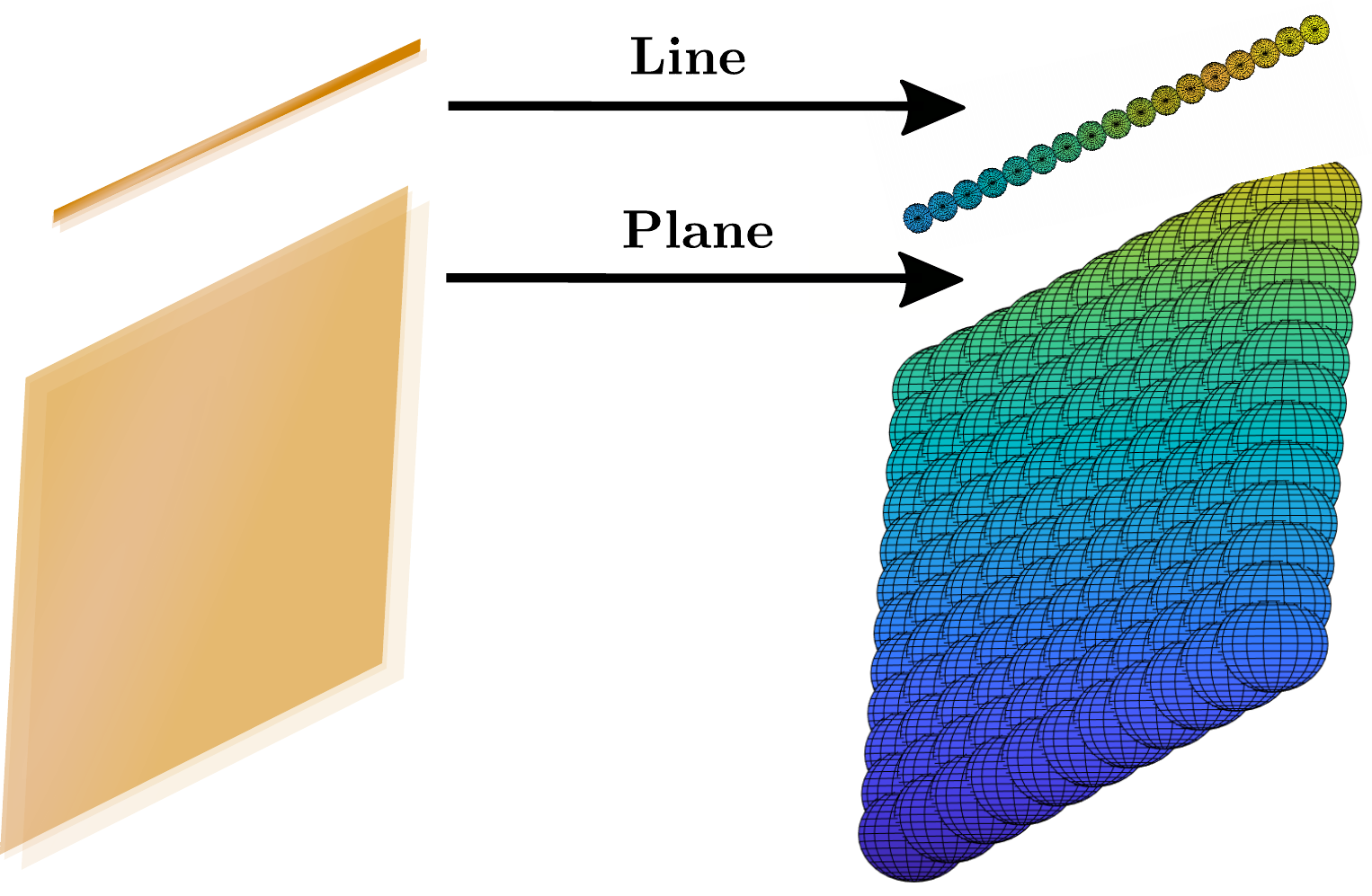}
    \caption{Spherization of Line and Plane}
    \label{fig:exp:plane scenario1}
\end{subfigure}
\begin{subfigure}{0.48\linewidth}
    \centering
    \includegraphics[width=0.6\linewidth]{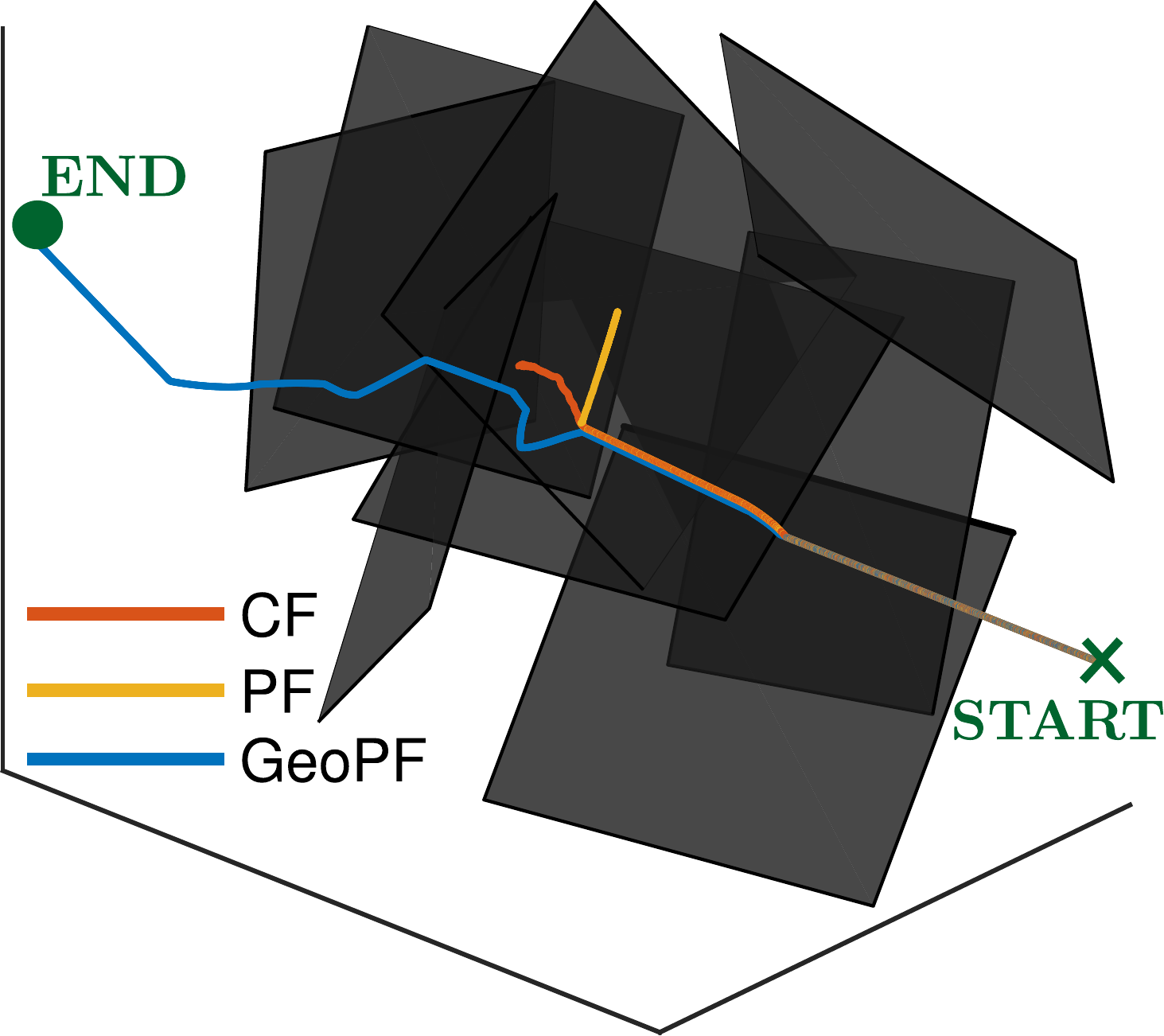}
    \caption{Trajectory in Plane Scenario}
    \label{fig:exp:plane scenario2}
\end{subfigure}
\caption{
Spherization in random Line and Plane Scenes. (b) 
shows how spherization scales objects from lines to planes, sharply increasing computation time. 
(c) GeoPF finds feasible paths,  
CF and PF are trapped in local minima, 
yielding 
lower success rates.}
\label{fig:exp:line spherization}
\end{figure}

\section{Quantitative Analysis and Experiments}

This section presents an extensive evaluation of GeoPF 
with two main goals,
(i) to quantitatively assess its performance%
%
\footnote{Key performance 
include (i) success rate—reaching the goal w/o collision, (ii) computational time; (iii)-(iv) path length and min. distance to obstacles. }
against 
state-of-the-art field-based planners across a variety of randomized scenes,  
and to demonstrate GeoPF’s feasibility, 
reliability\revised{, and robustness} in real 
experiments with a Franka \revised{robot}. 
To ensure fairness, all experiments were run on the same hardware,\footnote{AMD Ryzen\hspace{-1pt} 7\hspace{-1pt} 5800H, 32GB\hspace{-1pt} DDR4-3200 under Ubuntu 20.04, gcc/c++\hspace{-1pt} 9.4.\negmedspace} 
with randomized seeds, obstacle counts and placements, to thoroughly test robustness. 

Unless otherwise noted, we fix the same gains for GeoPF across all trials, highlighting its minimal tuning requirement.

\subsection{\textbf{Quantitative Analysis}}
\label{sec:QuantitativeAnalysis}
\revised{We evaluate planners under $1000$ randomized cases per scene} within a workspace 
$x,y,z {\in} [{-}0.2,0.2]$~m with initial position \revised{$[0, 1, 0]$} and goal $[0, -1, 0]$. State-of-the-art Baseline planners.  
%
%
\begin{itemize}
    \item     \textit{Potential Field} (PF): A standard artificial potential field with point- or sphere-based obstacle approximations; 
    \item \textit{Circular Field} (CF): A force-based variant exploiting circulatory vector fields along spherical obstacles~\cite{laha2021reactive}.
\end{itemize}
\revised{Tuning and spherization values were optimized to improve PF and CF results, under the following designed scenes. }
\begin{itemize}
    \item \textit{Line obstacles:} randomly generated lines 
    (Fig.~\ref{fig:exp:line spherization}) classified as  
    as “\textit{easy}” (5–10 lines) or “\textit{hard}” (10–50 lines);   
    \item \revised{\textit{Plane obstacles:}} randomly generated planes, classified as as \textit{easy} (2-8 planes) or \textit{hard} (10-40 planes). We also consider “\textit{longer}” scenarios (changing the goal position);
    \item \textit{Maze-Scenario}: designed over classic planning task, we compared the methods over different maze conditions. 
\end{itemize}
The results with key performance metrics
\footnote{
The minimum distance is the smallest value across the trajectory, while the average takes all the robot-obstacle distances through the whole trajectory. }   
%
are shown in Table~\ref{table::Field}.\hspace{-4pt} 

\noindent\textbf{Discussion and Results}

For \textit{Line Obstacle Scenarios}, 
\removed{both} 
the CF and PF require approximating each 
\removed{line}  
segment with small spheres \removed{as seen in} 
(Fig.~\ref{fig:exp:line spherization}). While small spheres ($r^{SP} = 0.01$) increase accuracy, they \revised{greatly increase the number of obstacle primitives and computation time.} 
Larger spheres $r^{SP} = 0.05$ reduce \minorrev{complexity} \removed{the count} but degrade fidelity.\footnote{%
Since PF is not as competitive as CF~\cite{laha2023predictive}, we only consider  $r^{SP} {=} 0.01$.
} 
%
\revised{GeoPf used a fixed $k^{Geo}=0.1$, across all tasks.} 
PF and CF mainly used $k^{attr}=1$, and $k^{attr}=0.1$ occasionally improving performance.
For each task, we performed a limited parameter search to identify optimal values for PF and CF.

\removed{The results, in}    
Table.~\ref{table::Field} \minorrev{shows GeoPF}
\removed{, show that GeoPF}   
consistently obtains higher success rates \minorrev{and up to $8\times$ faster computation.}
\removed{with shorter computational times (up to $8\times$ faster).} 
This is 
\revised{thanks not only to its closed-form distance and force computations, but also to its ability to exploit the geometric structure of lines and planes, differentiating and modulating repulsive forces based on interactions with faces, edges, or vertices. Instead, existing methods rely solely on the closes-contact point---except for CF, which uses sphere-based modulation but introduces significant overhead, especially in cluttered scenes.} 
\begin{figure}[t]
\centering
\includegraphics[width=1\linewidth]{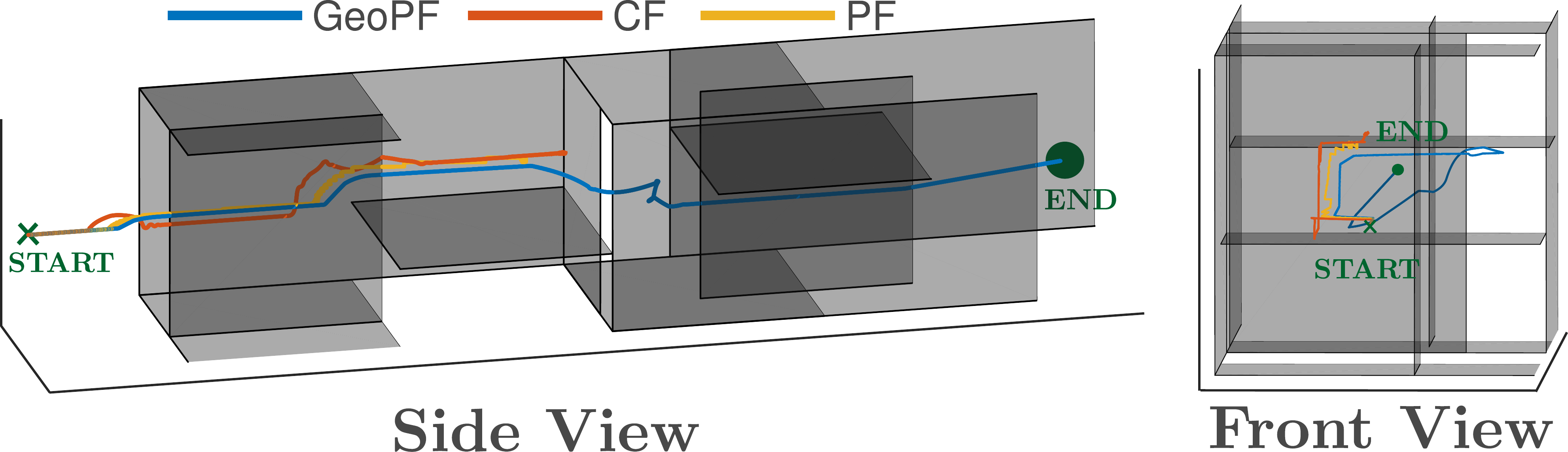}
\caption{Maze Task. The robot should find a feasible path with avoiding the plane and go through the tunnel. }
\label{fig::exp::tunnel}
\end{figure}

\begin{figure}[t]
		\centering
		\includegraphics[width=\linewidth]{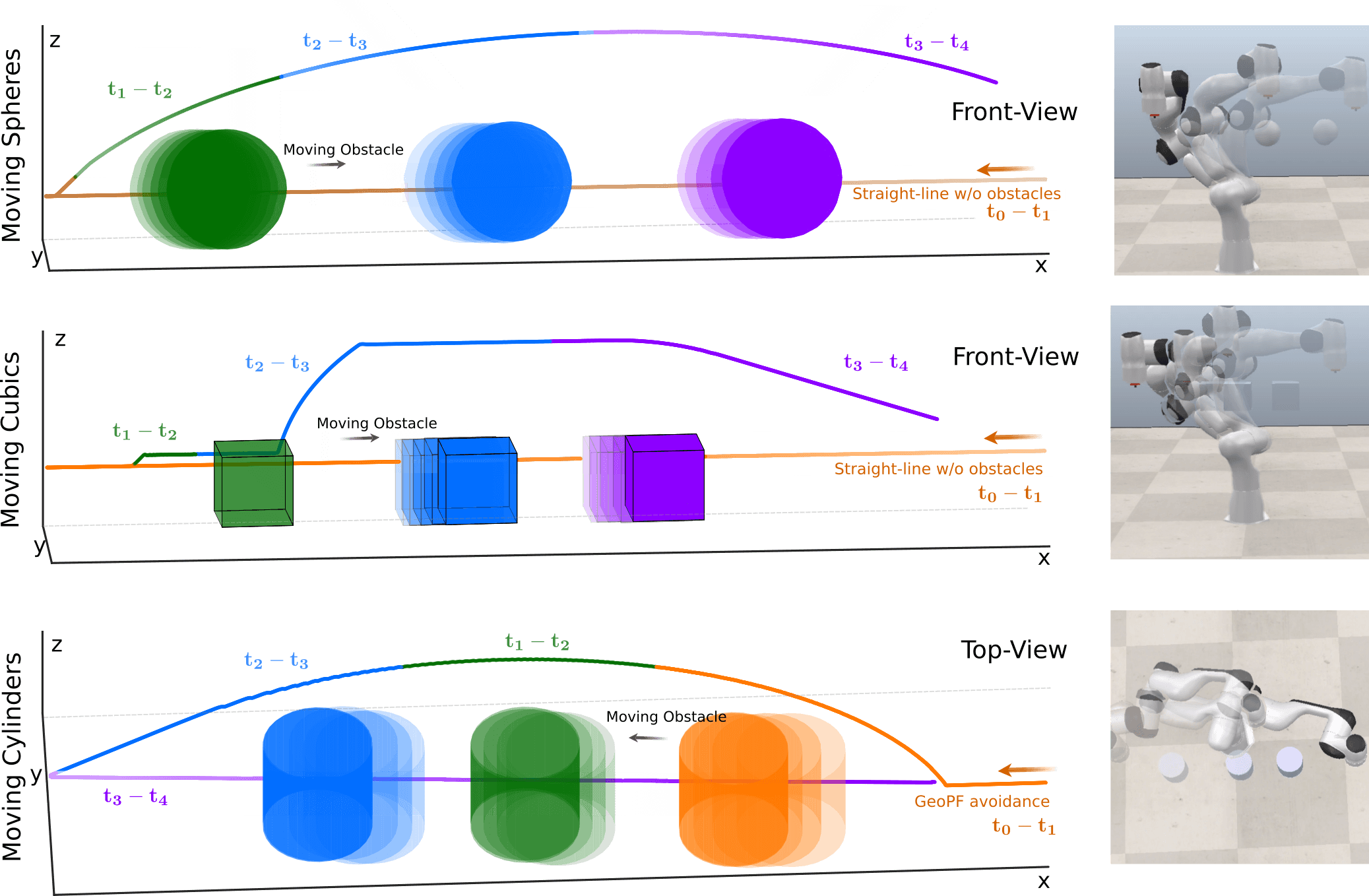}
\caption{
\revised{Dynamic obstacle avoidance with different primitive-specific modulation, 
in response to 
sphere (a), cube (b), and cylinder (c).
}}
\label{fig:dynamic:primitives}
\end{figure}

\begin{table*}[t]
\centering
\caption{Quantitative Analysis (\textcolor{blue!70!black}{mean}/\textcolor{blue!70!gray!70}{Std}) 
between CF, PF, and GeoPF planners. 
Key metrics: 
\revised{\textbf{N.Obs.Pr.} (number of obstacle primitives)}, 
\textbf{SR} (success rate), 
\textbf{Step} (iterations), 
\textbf{CT/Step (s)} (computation time per step), 
\textbf{PL} (path length), 
\textbf{Min} and \textbf{Av}. \textbf{Dist.} (distance to obstacles).
}
%
%
\label{table::Field}
\setlength{\tabcolsep}{9pt}
\begin{tabular}{@{}lcccccccccc@{}}
\toprule
&
M. ($r^{SP}, k^{attr}$) &
N.Obs.Pr. &
SR &
\multicolumn{1}{c}{Step} &
\multicolumn{1}{c}{CT/Step (ms)} &
PL (m) &
\hspace{-10pt}Min. Dist. (m)&
\hspace{-10pt}Av. Dist. (m)\\
\midrule

\multirow{4}{*}{\makecell{Line\\(easy)}} 
& \textcolor{blue!70!black}{GeoPF}
  & \textbf{\textcolor{blue!70!black}{10}/\textcolor{blue!70!gray!70}{2.8}}
  & \textcolor{blue!70!black}{\textbf{100}\%}
  & \textcolor{blue!70!black}{4909}/\textcolor{blue!70!gray!70}{62.83}
  & \textbf{\ctperstepms{0.32}{0.08}{4909}{62.83}}
  & \textcolor{blue!70!black}{2.03}/\textcolor{blue!70!gray!70}{0.03}
  & \textcolor{blue!70!black}{.046}/\textcolor{blue!70!gray!70}{.001}
  & \textbf{\textcolor{blue!70!black}{.510}/\textcolor{blue!70!gray!70}{.001}} \\

& CF ($0.01,1$)
  & 135/\textcolor{gray}{77.46}
  & \textbf{100}\%
  & 5044/\textcolor{gray}{81.69}
  & \ctperstepms{1.16}{0.3464}{5044}{81.69}
  & 2.210/\textcolor{gray}{0.044}
  & .061/\textcolor{gray}{.001}
  & .504/\textcolor{gray}{.001} \\

& CF ($0.05,1$)
  & 31/\textcolor{gray}{17.78}
  & 99.4\%
  & 5140/\textcolor{gray}{118}
  & \ctperstepms{0.57}{0.155}{5140}{118}
  & 2.175/\textcolor{gray}{0.158}
  & .049/\textcolor{gray}{.001}
  & .477/\textcolor{gray}{.001} \\

& PF ($0.01,1$)
  & 135/\textcolor{gray}{77.46}
  & 99.2\%
  & \textbf{4860/\textcolor{gray}{14.42}}
  & \ctperstepms{1.10}{0.326}{4860}{14.42}
  & \textbf{1.998/\textcolor{gray}{0.005}}
  & .017/\textcolor{gray}{.001}
  & .505/\textcolor{gray}{.001} \\

\midrule
\multirow{4}{*}{\makecell{Line\\(hard)}}
& \textcolor{blue!70!black}{GeoPF}
  & \textbf{\textcolor{blue!70!black}{30}/\textcolor{blue!70!gray!70}{11.62}}
  & \textbf{\textcolor{blue!70!black}{100\%}}
  & \textbf{\textcolor{blue!70!black}{5025}/\textcolor{blue!70!gray!70}{86.04}}
  & \textbf{\ctperstepms{0.94}{0.36}{5025}{86.04}}
  & \textbf{\textcolor{blue!70!black}{2.097}/\textcolor{blue!70!gray!70}{0.055}}
  & \textcolor{blue!70!black}{.032}/\textcolor{blue!70!gray!70}{.001}
  & \textcolor{blue!70!black}{.505}/\textcolor{blue!70!gray!70}{.001} \\

& CF ($0.01,0.1$)
  & 462/\textcolor{gray}{316}
  & 100\%
  & 11258/\textcolor{gray}{195}
  & \ctperstepms{7.65}{2.986}{11258}{195}
  & 2.287/\textcolor{gray}{0.105}
  & .051/\textcolor{gray}{.001}
  & \textbf{.686/\textcolor{gray}{.001}} \\

& CF ($0.05, 1$)
  & 101/\textcolor{gray}{70.71}
  & 88.7\%
  & 5270/\textcolor{gray}{126.73}
  & \ctperstepms{1.55}{0.599}{5270}{126.73}
  & 2.287/\textcolor{gray}{0.21}
  & .031/\textcolor{gray}{.001}
  & .485/\textcolor{gray}{.001} \\

& PF ($0.01, 0.1$)
  & 462/\textcolor{gray}{316}
  & 47\%
  & 11408/\textcolor{gray}{431}
  & \ctperstepms{6.96}{3.13}{11408}{431}
  & 2.271/\textcolor{gray}{0.10}
  & .078/\textcolor{gray}{.001}
  & .683/\textcolor{gray}{.001} \\

\bottomrule
\toprule
&
\textbf{M.} ($r^{SP}, k^{SP}$) &
\textcolor{blue}{N.Obs.Pr.} &
SR &
\multicolumn{1}{c}{Step} &
\multicolumn{1}{c}{CT/Step (ms)} &
PL (m) &
\hspace{-10pt}Min. Dist. (m)&
\hspace{-10pt}Av. Dist. (m)\\
\midrule
\multirow{4}{*}{\makecell{Plane\\(easy)}}
& \textcolor{blue!70!black}{GeoPF}
  & \textbf{\textcolor{blue!70!black}{5}/\textcolor{blue!70!gray!70}{2.236}}
  & \textbf{\textcolor{blue!70!black}{97\%}}
  & \textcolor{blue!70!black}{5362}/\textcolor{blue!70!gray!70}{528}
  & \textbf{\ctperstepms{0.84}{0.356}{5362}{528}}
  & \textcolor{blue!70!black}{2.29}/\textcolor{blue!70!gray!70}{0.302}
  & \textcolor{blue!70!black}{.037}/\textcolor{blue!70!gray!70}{.002}
  &\textbf{ \textcolor{blue!70!black}{.464}/\textcolor{blue!70!gray!70}{.002}} \\

& CF (0.01, 100)
  & 1e4/\textcolor{gray}{1e5}
  & 74.7\%
  & 5211/\textcolor{gray}{300}
  & \ctperstepms{88.8}{37.53}{5211}{300}
  & 2.15/\textcolor{gray}{0.145}
  & .080/\textcolor{gray}{.001}
  & .450/\textcolor{gray}{.001} \\

& PF (0.01, 100)
  & 1e4/\textcolor{gray}{1e4}
  & 54.5\%
  & 6324/\textcolor{gray}{1343.5}
  & \ctperstepms{111.2}{63.45}{6324}{1343.5}
  & 3.1/\textcolor{gray}{0.95}
  & .110/\textcolor{gray}{.003}
  & .440/\textcolor{gray}{.003} \\

& CF (0.05, 10)
  & 584/\textcolor{gray}{547}
  & 31.5\%
  & \textbf{4881/\textcolor{gray}{73.03}}
  & \ctperstepms{3.54}{1.49}{4881}{73.03}
  &\textbf{ 2.01/\textcolor{gray}{0.03}}
  & .029/\textcolor{gray}{.002}
  & .418/\textcolor{gray}{.002} \\

\midrule
\multirow{4}{*}{\makecell{\hspace{-28pt} Plane\\(easy-longer)}}
& \textcolor{blue!70!black}{GeoPF}
  & \textbf{\textcolor{blue!70!black}{5}/\textcolor{blue!70!gray!70}{2.236}}
  & \textbf{\textcolor{blue!70!black}{97.9\%}}
  & \textbf{\textcolor{blue!70!black}{5525}/\textcolor{blue!70!gray!70}{592}}
  & \textbf{\ctperstepms{0.87}{0.373}{5525}{592}}
  & \textcolor{blue!70!black}{2.33}/\textcolor{blue!70!gray!70}{0.232}
  & \textcolor{blue!70!black}{.039}/\textcolor{blue!70!gray!70}{.042}
  & \textbf{\textcolor{blue!70!black}{.480}/\textcolor{blue!70!gray!70}{.045}} \\

& CF (0.01, 100)
  & 1e4/\textcolor{gray}{1e4}
  & 78.9\%
  & 5437/\textcolor{gray}{308}
  & \ctperstepms{99.1}{41.43}{5437}{308}
  & 2.25/\textcolor{gray}{0.13}
  & .069/\textcolor{gray}{.002}
  & .453/\textcolor{gray}{002} \\

& PF (0.01, 100)
  & 1e4/\textcolor{gray}{1e4}
  & 69.4\%
  & 5280/\textcolor{gray}{360}
  & \ctperstepms{89.7}{41.94}{5280}{360}
  & \textbf{2.169/\textcolor{gray}{0.148}}
  & .101/\textcolor{gray}{.001}
  & .452/\textcolor{gray}{.002} \\

& PF (0.05, 100)
  & 584/\textcolor{gray}{548}
  & 60.2\%
  & 5281/\textcolor{gray}{374}
  & \ctperstepms{4.10}{1.967}{5281}{374}
  & 2.170/\textcolor{gray}{0.152}
  & .101/\textcolor{gray}{.001}
  & .424/\textcolor{gray}{.001} \\

\midrule
\multirow{4}{*}{\makecell{Plane\\(hard)}}
& \textcolor{blue!70!black}{GeoPF}
  & \textbf{\textcolor{blue!70!black}{25}/\textcolor{blue!70!gray!70}{10.72}}
  & \textbf{\textcolor{blue!70!black}{59\%}}
  & \textbf{\textcolor{blue!70!black}{5297}/\textcolor{blue!70!gray!70}{881}}
  &\textbf{ \ctperstepms{4.27}{2.18}{5297}{881}}
  & \textcolor{blue!70!black}{2.641}/\textcolor{blue!70!gray!70}{0.557}
  & \textcolor{blue!70!black}{.015}/\textcolor{blue!70!gray!70}{.001}
  & \textcolor{blue!70!black}{.589}/\textcolor{blue!70!gray!70}{.002} \\

& CF (0.01, 100)
  & 6e4/\textcolor{gray}{54770}
  & 46\%
  & 5923/\textcolor{gray}{547.7}
  & \ctperstepms{489}{244.9}{5923}{547.7}
  & 2.49/\textcolor{gray}{0.293}
  & .044/\textcolor{gray}{.001}
  & .585/\textcolor{gray}{.002} \\

& CF (0.05, 10)
  & 3e3/\textcolor{gray}{7746}
  & 29\%
  & 5521/\textcolor{gray}{379.4}
  & \ctperstepms{19.3}{9.62}{5521}{379.4}
  & \textbf{2.273/\textcolor{gray}{0.16}}
  & .040/\textcolor{gray}{.001}
  & .539/\textcolor{gray}{.003} \\

& PF (0.05, 1)
  & 3e3/\textcolor{gray}{7746}
  & 11.2\%
  & 13533/\textcolor{gray}{2683}
  & \ctperstepms{39.93}{26.01}{13533}{2683}
  & 3.136/\textcolor{gray}{1.039}
  & .099/\textcolor{gray}{.001}
  & \textbf{.632/\textcolor{gray}{.001}} \\

\midrule
\multirow{4}{*}{\makecell{\hspace{-28pt} Plane\\(hard-longer)}}
& \textcolor{blue!70!black}{GeoPF}
  & \textbf{\textcolor{blue!70!black}{25}/\textcolor{blue!70!gray!70}{10.72}}
  & \textbf{\textcolor{blue!70!black}{62.8\%}}
  & \textcolor{blue!70!black}{6108}/\textcolor{blue!70!gray!70}{632.5}
  & \textbf{\ctperstepms{4.39}{2.02}{6108}{632.5}}
  & \textcolor{blue!70!black}{2.72}/\textcolor{blue!70!gray!70}{0.528}
  & \textcolor{blue!70!black}{.011}/\textcolor{blue!70!gray!70}{.001}
  &\textbf{ \textcolor{blue!70!black}{.617}/\textcolor{blue!70!gray!70}{.004}} \\

& CF (0.01, 100)
  & 6e4/\textcolor{gray}{54772}
  & 49.4\%
  & 5887/\textcolor{gray}{547.7}
  & \ctperstepms{489.4}{244.9}{5887}{547.7}
  & 2.48/\textcolor{gray}{0.29}
  & .046/\textcolor{gray}{.001}
  & .585/\textcolor{gray}{.002} \\

& CF (0.01, 1)
  & 6e4/\textcolor{gray}{54772}
  & 17.4\%
  &\textbf{ 5038/\textcolor{gray}{63.88}}
  & \ctperstepms{454}{221.36}{5038}{63.88}
  & \textbf{2.09/\textcolor{gray}{0.048}}
  & .010/\textcolor{gray}{.001}
  & .582/\textcolor{gray}{.001} \\

& PF (0.01, 100)
  & 6e4/\textcolor{gray}{54772}
  & 13.6\%
  & 6833/\textcolor{gray}{1483}
  & \ctperstepms{417}{173.2}{6833}{1483}
  & 3.45/\textcolor{gray}{1.068}
  & .099/\textcolor{gray}{.001}
  & .572/\textcolor{gray}{.001} \\

\midrule
\bottomrule
\end{tabular}
\end{table*}

\begin{table}[t]
\centering
\caption{GeoPF performance in complex and dynamic scenarios. Average (black) and standard deviation (gray) shown. \textbf{CT/S}. represents \textbf{CT/Step}. \textbf{Min.D.} represents \textbf{Min.Dist.}}
\label{table::Geo_compact}
\begin{tabular}{@{}lcccc@{}}
\toprule
\textbf{Scenario} & \textbf{N.Obs.Pr.} & \textbf{SR} & \textbf{CT/S. (ms)} & \textbf{\hspace{-2pt}Min.\hspace{-1pt} D. \hspace{-2pt}(m)} \\
\midrule
Complex & \textcolor{black}{$12$}/\textcolor{gray}{$2.45$} & \textcolor{black}{93.7\%} & \textbf{\textcolor{black}{$0.62$}}/\textcolor{gray}{$0.37$} & \textcolor{black}{$0.032$}/\textcolor{gray}{$0.03$} \\

Dynamic (easy) & \textcolor{black}{$\frac{15}{2}(\frac{5}{2})$}/\textcolor{gray}{$1.1(1.05)$} & \textcolor{black}{94.9\%} & \textbf{\textcolor{black}{$0.48$}}/\textcolor{gray}{$0.07$} & \textcolor{black}{$0.026$}/\textcolor{gray}{$0.02$} \\

Dynamic (hard) & \textcolor{black}{$15(7)$}/\textcolor{gray}{$1.73(1)$} & \textcolor{black}{83.2\%} & \textbf{\textcolor{black}{$0.89$}}/\textcolor{gray}{$0.25$} & \textcolor{black}{$0.017$}/\textcolor{gray}{$0.02$} \\
\bottomrule
\end{tabular}
\end{table}

For \textit{Plane-Obstacle Scenarios},\footnote{\revised{New parameter search was required for PF and CF, as prior values (used successfully in line-based scenes) consistently failed. We analyzed repulsive coefficients  $k^{SP}=1,10,100$. GeoPF kept the same parameters.}} 
spherization led to massive sphere counts and excessive computational times. Except in simple cases, PF and CF could not produce feasible solutions within real-time constraints, as shown in Table~II. In contrast, GeoPF reduced computational time by up to $100\times$, achieving real-time (below 1~ms)  in all cases while maintaining higher success rates even in dense and cluttered scenarios. 
\removed{we conducted a new parameter search to identify optimal values for PF and CF, as the previously used parameters consistently led to failure. In this evaluation, we analyzed different repulsive coefficients for PF and CF, $k^{SP}=1,10,100$.  GeoPF uses the same parameters as in the line task, with $k^{attr}=1$, $k^{Geo}=0.1$.

The results indicate that plane spherization generates a significantly large number of spheres, resulting in an extremely high computational time. 
Indeed, except for simple line-based scenes, CF and PF planners fail to generate a  feasible solution within the real-time framework, as seen in the CT/Step (ms) column in Table~\ref{table::Field}.  

In contrast, 
GeoPF drastically lowers the total computational time---up to $100\times$ faster in almost all cases---thanks to direct distance-to-plane calculations. It is also able to keep the real-time performance (computational below 1~ms) for all cases,11 while maintaining a higher success rate in more cluttered arrangements. 
}



\removed{
The classic \textit{Maze Task} scenario, where the robot must find a feasible path across multiple planes positioned either vertically or horizontally, is considered a standard trap case.  
As shown in Fig.~\ref{fig::exp::tunnel}, while CF and PF fail at the centre point in front of the middle plane, GeoPF successfully navigates through the maze. This result reinforces the findings in Table~\ref{table::Field}  which show that CF and PF exhibit significantly lower success in complex scenarios.}

\minorrev{In the challenging \textit{Maze Task}, considered a standard trap case for reactive planners with multiple vertical and horizontal planes (Fig.~\ref{fig::exp::tunnel}), } 
CF and PF fail at the centre point in front of the middle plane, GeoPF successfully navigates through the maze, aligning with results from Table~\ref{table::Field}.

\noindent \textbf{Complex and Dynamic Scenes}

We 
\removed{additionally tested GeoPF}  
\minorrev{further evaluated}  
in challenging environments composed of multiple \removed{geometric }primitives---including lines, planes, cubes, and cylinders\minorrev{, in cluttered and dynamic configurations.}\removed{---all placed or moving in random configurations. In these trials, only GeoPF was evaluated, as our goal was to demonstrate its versatility in cluttered, complex, and dynamic scenarios. 
}
\removed{Specifically, each}
\minorrev{Each} 
composite scene randomly \removed{selected}\minorrev{included} 5–10 lines, 2–5 planes, and 2–3 cubes or cylinders distributed around the workspace. 
\removed{Some obstacles were static, while others followed randomized trajectories (e.g., drifting planes or cylinders).
In Table~\ref{table::Field}, the notation \textcolor{blue}{\textbf{N.Obs.Pr.}} is represented as $a(b)$, where $a$ denotes the total \textcolor{blue}{number of obstacle primitives}, while (b) the dynamic ones.
}

\removed{Throughout these tests, GeoPF maintained} 
\minorrev{GeoPF consistently achieved} high success rates without requiring special tuning or re-initialization, see Table~\ref{table::Geo_compact}.
\revised{Fig.~\ref{fig:dynamic:primitives} further shows the dynamic response modulated per primitive in a back-and-forth scene (with dynamic obstacles appearing in the middle of an originally straight line goal).}
The same $k$ value parameters were used \revised{throughout this paper}. 
\revised{These results highlight GeoPF’s flexibility, real-time performance, and stable behavior in complex, dynamic scenes.}

\removed{
These results confirm the flexibility and easiness to use, as well as the reliability of GeoPF which preserves its real-time efficiency in more complex scenes with a stable and reliable behaviour.
}

\removed{
\begin{figure}[t]
\centering
\includegraphics[width=1\linewidth]{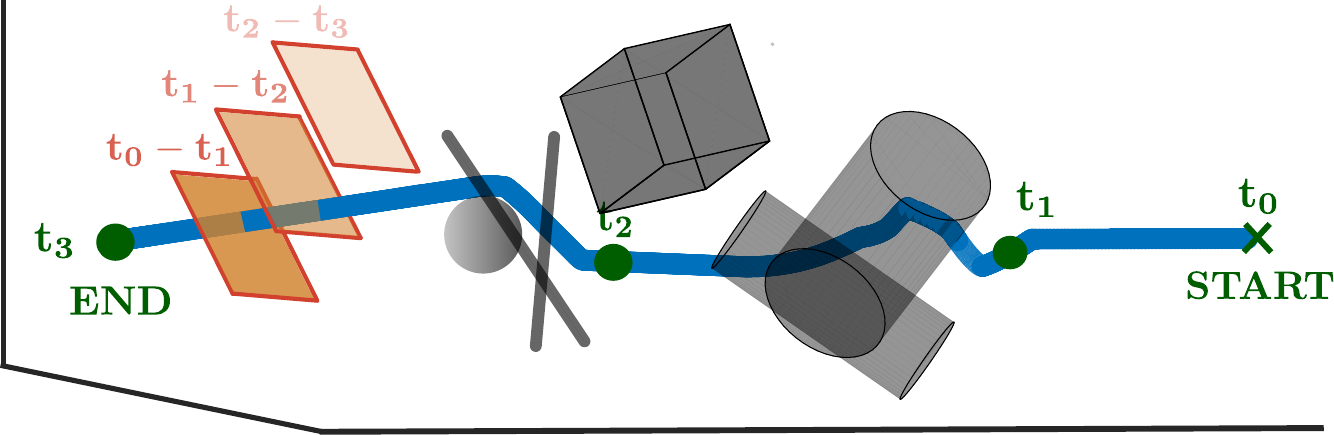}
\caption{
{Dynamic scenarios with plane moving upwards during execution.}
}
\label{exp::fig::dynamic}
\end{figure}
}

\subsection{\textbf{Real Robot Experiment}}
\label{sec:Experiment}
To validate GeoPF's \removed{the} practical viability, 
we deployed it on a 7-DoF Franka Robot.  
End-effector velocities were generated by mapping the 
GeoPF forces into joint-space 
via Jacobian pseudoinverse. 
Obstacles were tracked via a motion-capture system (Nokov Mars) and fitted to corresponding geometric primitives. GeoPF was tested 
across \revised{four}
representative tasks. 

\begin{figure}[t]
\centering
\begin{subfigure}{0.49\linewidth}
    \centering
    \includegraphics[width=0.95\linewidth]{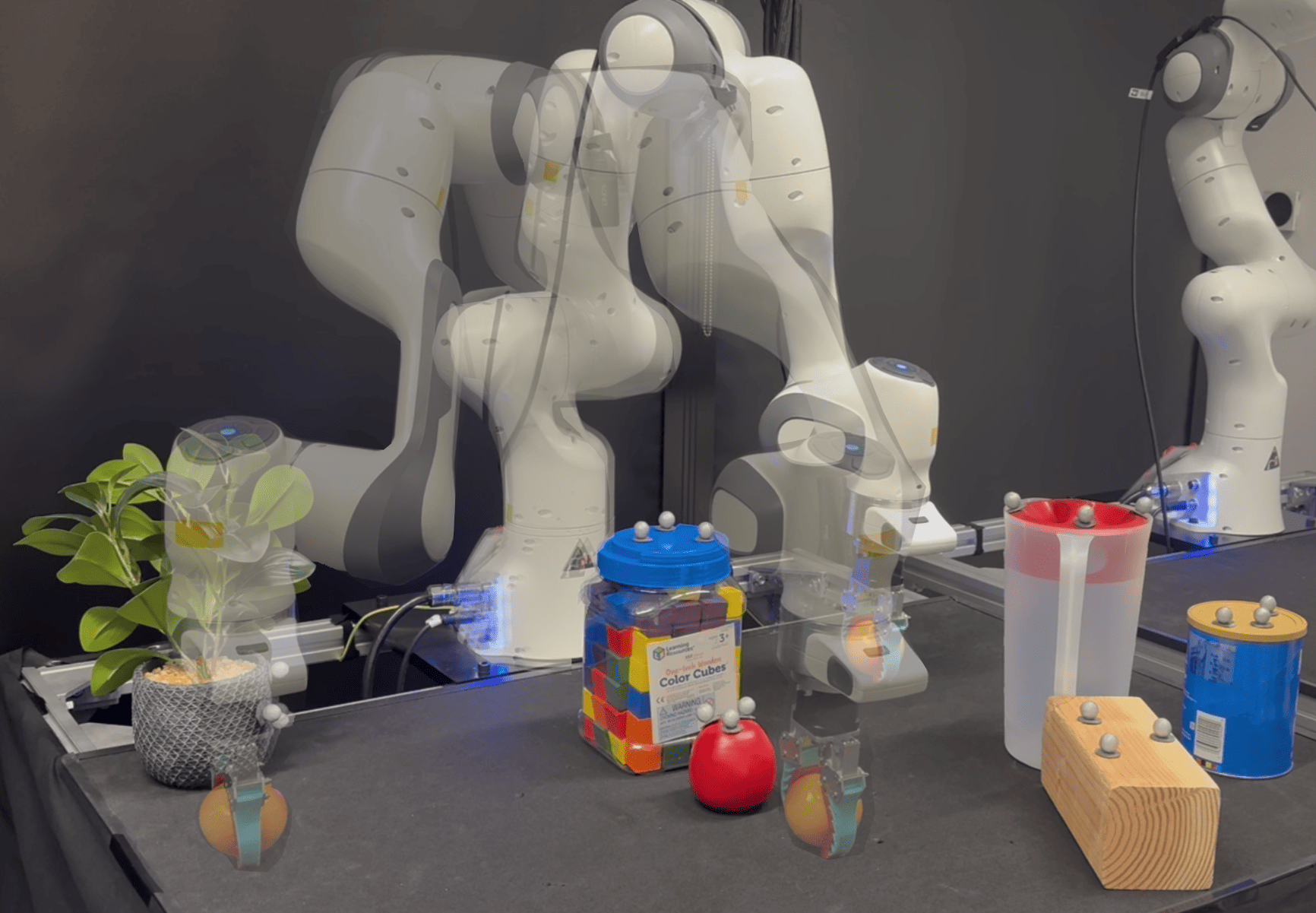}
    \caption{2D Task in Real Robot}
    \label{exp::fig::real1}
\end{subfigure}
\begin{subfigure}{0.49\linewidth}
    \centering
    \includegraphics[width=1\linewidth]{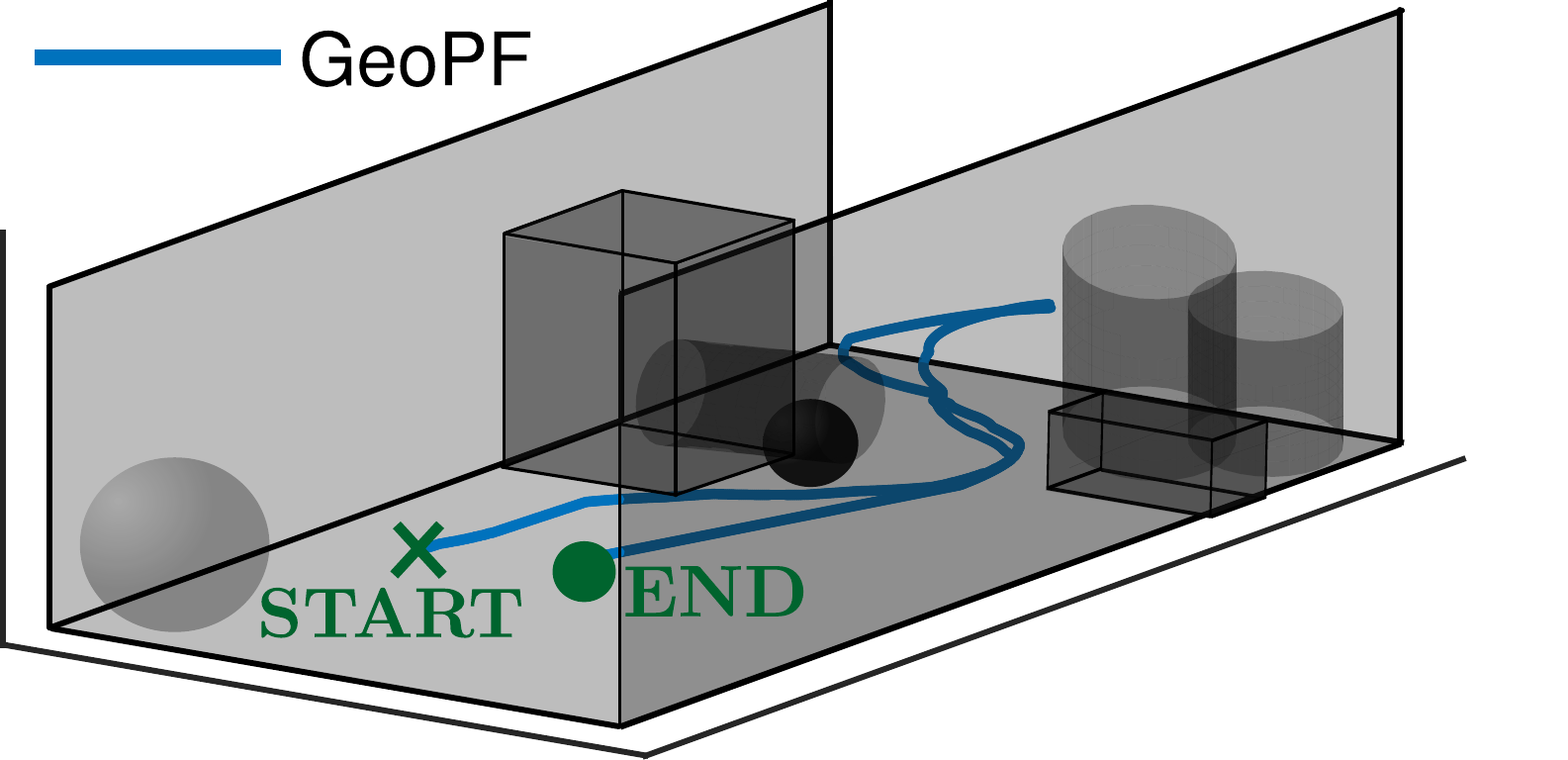}
    \caption{Trajectory for 2D Task}
    \label{exp::fig::traj1}
\end{subfigure}
\begin{subfigure}{0.49\linewidth}
    \centering
    \includegraphics[width=0.95\linewidth]{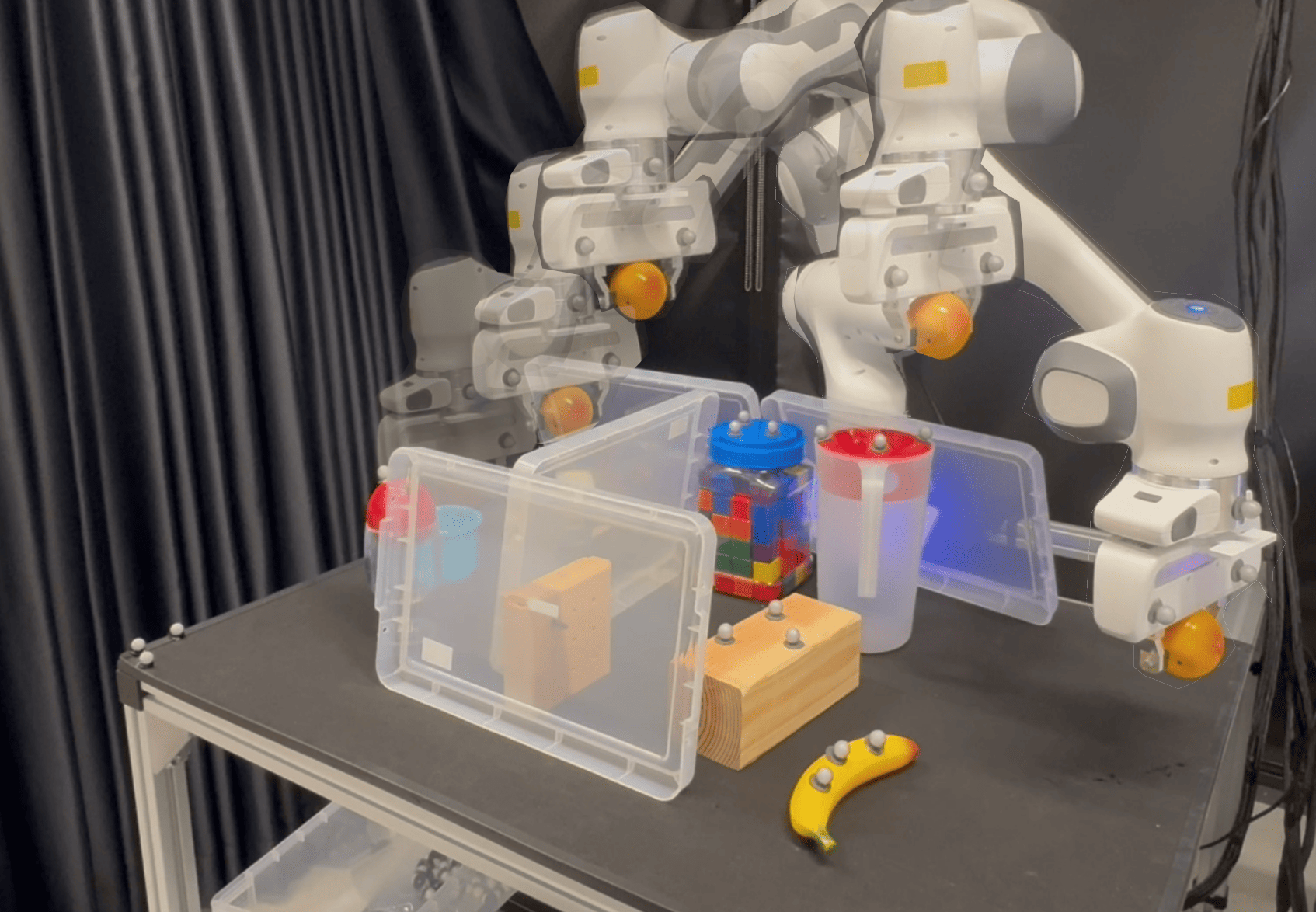}
    \caption{3D Task in Real Robot}
    \label{exp::fig::real2}
\end{subfigure}
\begin{subfigure}{0.49\linewidth}
    \centering
    \includegraphics[width=0.7\linewidth]{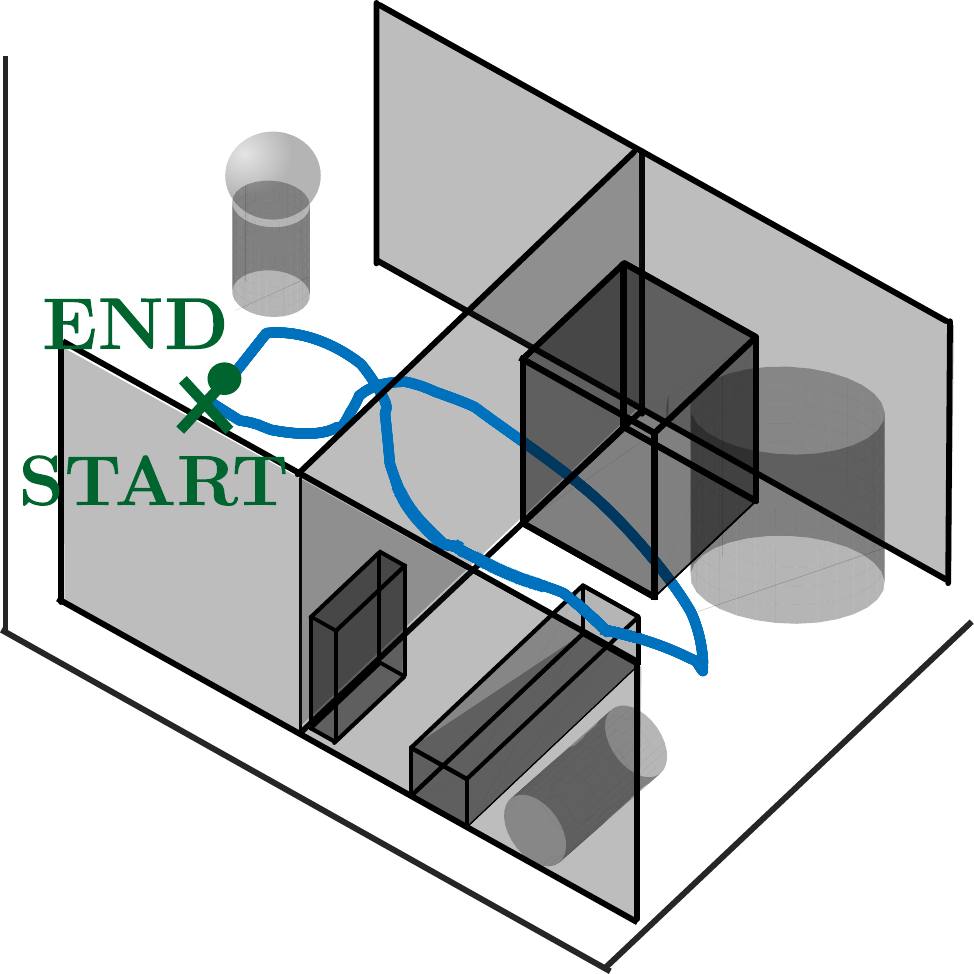}
    \caption{Trajectory for 3D Task}
    \label{exp::fig::traj1}
\end{subfigure}
\caption{Real world experiments. The robot goes from left 
to right (high-opacity) and then returns, see trajectories (b),(d). The \textit{upper figures} show a 2D tabletop task  with multiple obstacles, as lines, cubes, and cylinders. The workspace
is shown by planes, 
w.r.t. the robot's boundaries. 
The \textit{bottom figures} illustrate a 3D task with four lids representing walls that the robot must avoid during navigation. 
}  
\label{exp::fig::real}
\end{figure}

\begin{figure}[t]
\centering\vspace{-10pt}
\begin{subfigure}{0.24\linewidth}
		\centering
		\includegraphics[width=1\linewidth]{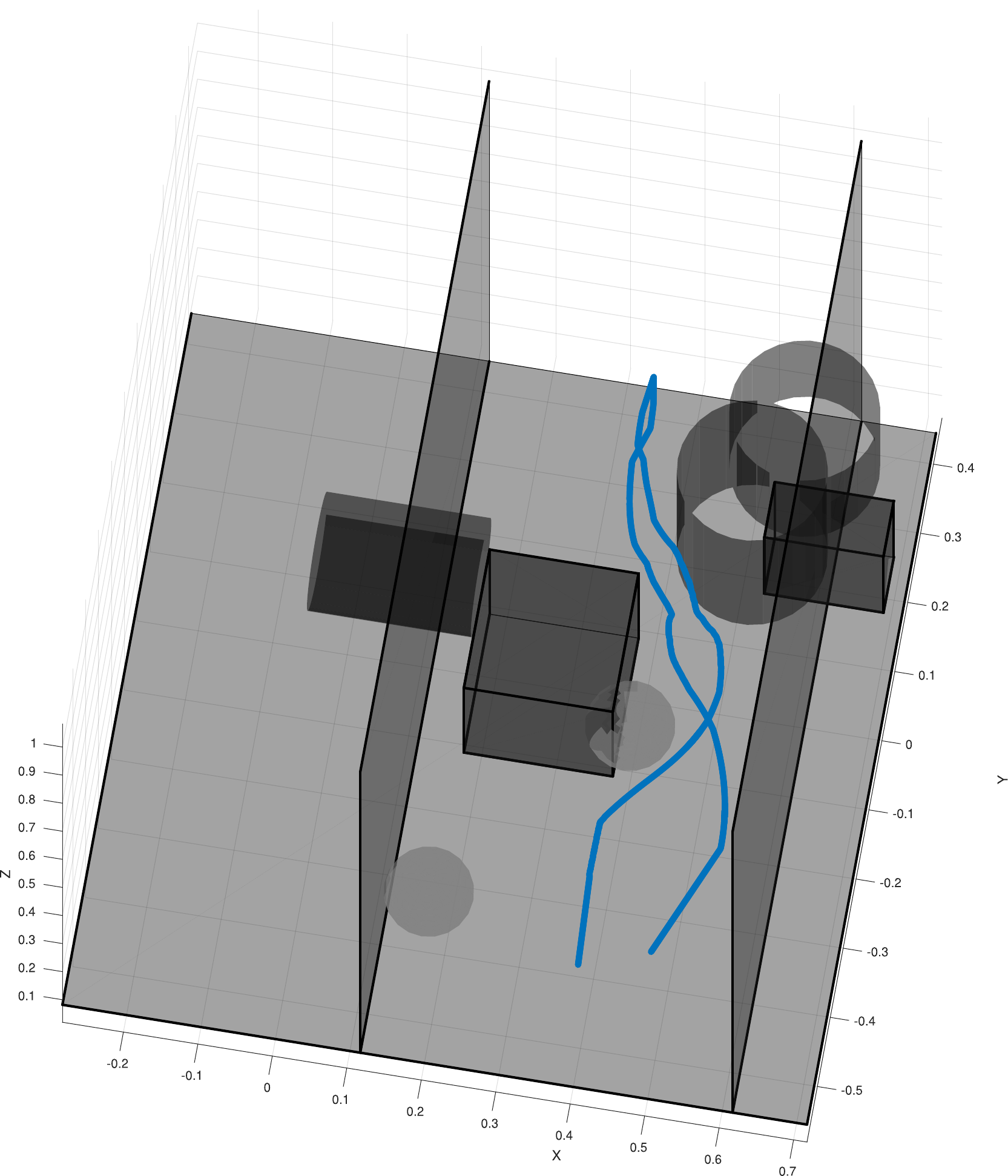}
        \vspace{-20pt}\caption{\hspace{30pt} }
            \label{fig:GeoPF}
	\end{subfigure}
\begin{subfigure}{0.24\linewidth}
		\centering
		\includegraphics[width=\linewidth]{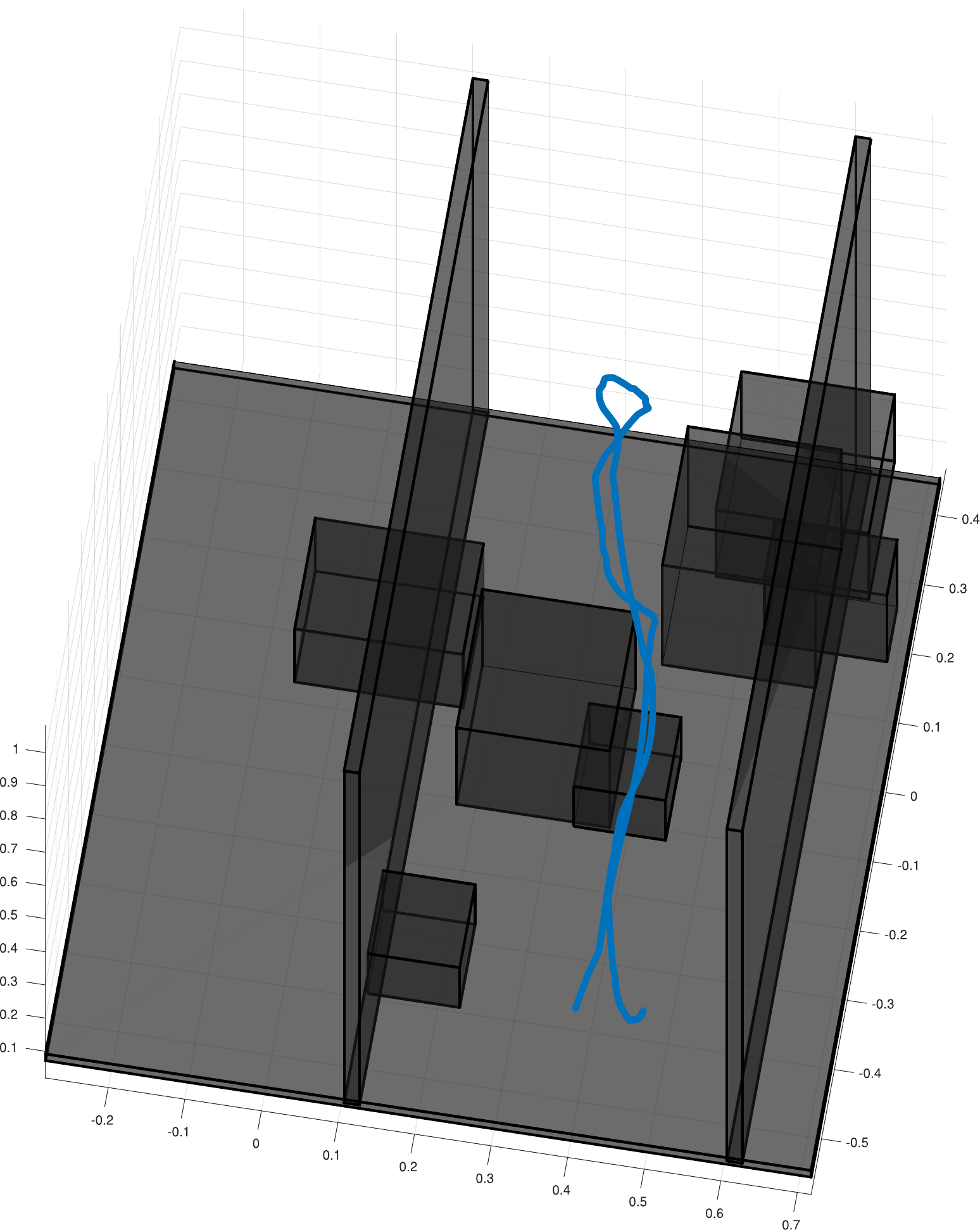}
        \vspace{-20pt}\caption{\hspace{30pt} }
            \label{fig:Cubic}
	\end{subfigure}
\begin{subfigure}{0.24\linewidth}
		\centering
		\includegraphics[width=\linewidth]{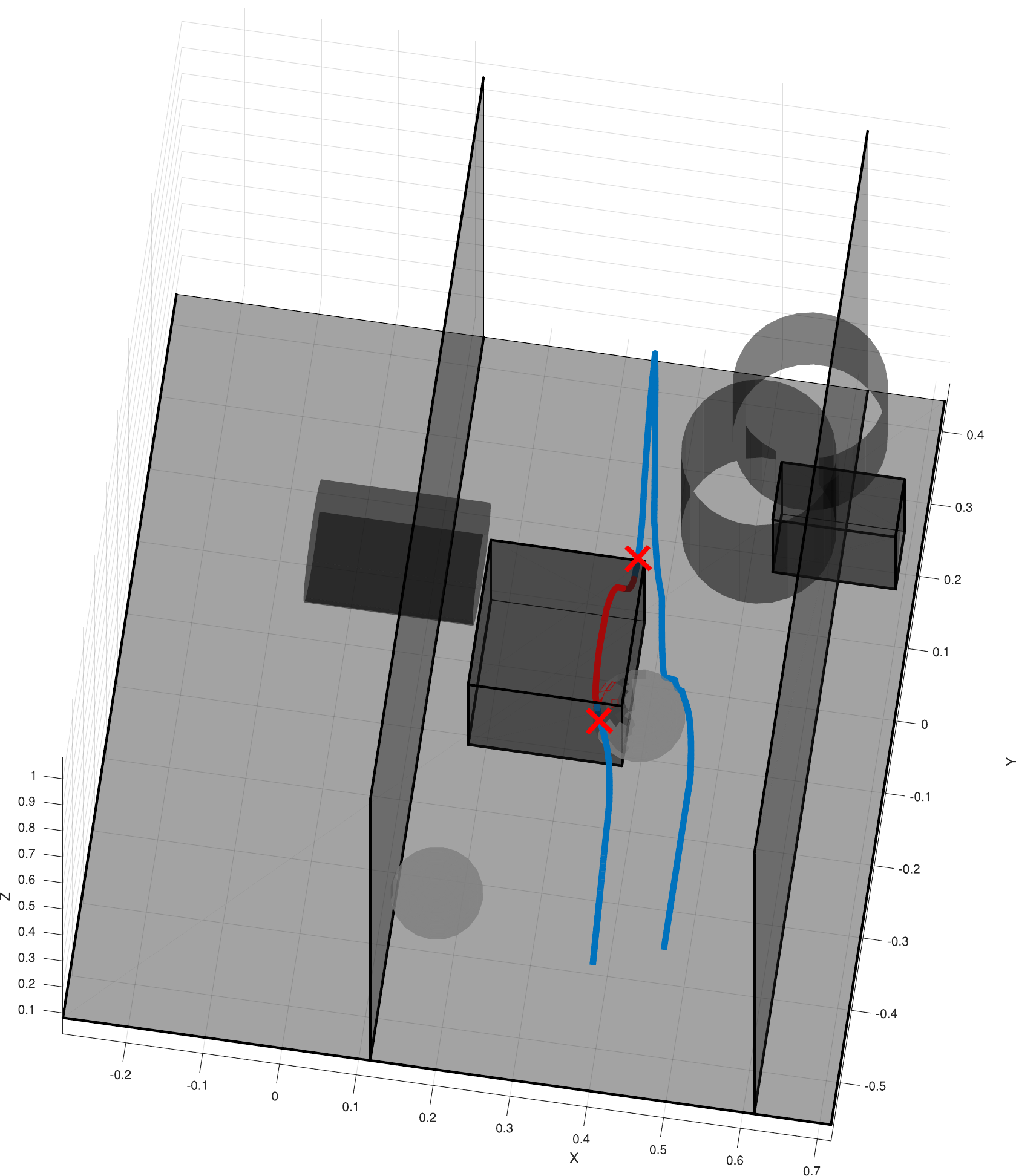}
         \vspace{-20pt}\caption{\hspace{30pt} }
            \label{fig:CF}
	\end{subfigure}
\begin{subfigure}{0.24\linewidth}
		\centering
		\includegraphics[width=\linewidth]{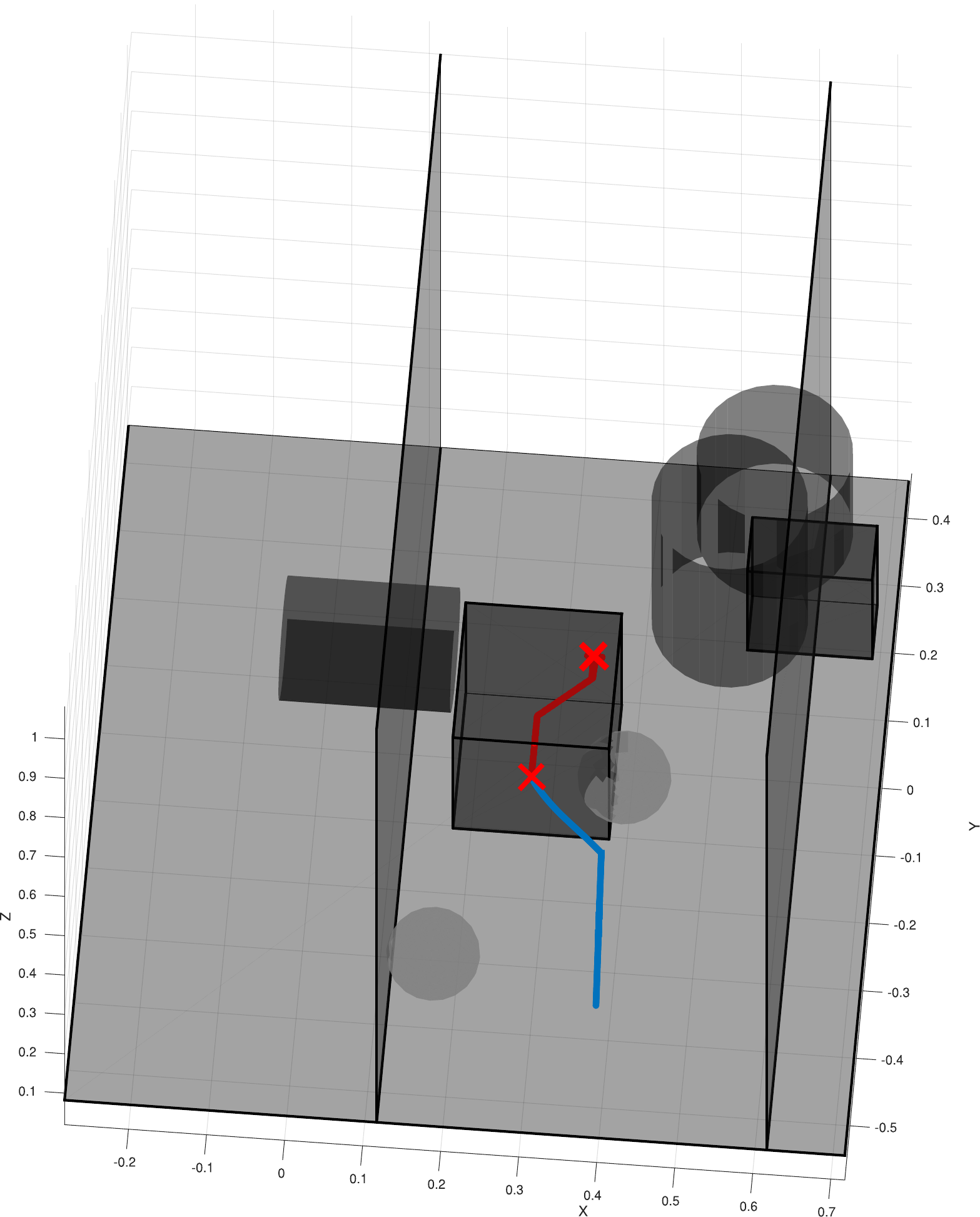}
         \vspace{-20pt}\caption{\hspace{30pt} }
            \label{fig:PF}
	\end{subfigure}
%
%
\caption{\revised{Robustness to primitive fitting---based on 2D Task (Fig.~\ref{exp::fig::real}(a)). 
\textbf{(a)} GeoPF with structured primitives-fitting produces high-quality, shape-aware trajectories, while CFs \textbf{(c)} and APFs \textbf{(d)} fail to generate a collision-free trajectory, despite being given full geometric-information. \textbf{(b)} GeoPF with only access to bounding boxes still finds a feasible path---even with less privileged information. The red lines indicate that robot goes into the obsctale.
%
}
}
\label{fig:real:robustness}
\end{figure}

First (Fig.~\ref{exp::fig::real1}), 
in a tabletop scenario,
GeoPF leveraged geometric primitives to produce 
smooth, collision-free trajectories under rigid (z-axis) constraints\revised{, avoiding randomly placed lines, cubes, and cylinders within plane-bounded workspaces.}  
\revised{
To further evaluate robustness to primitive shaping, we revised the same experiment, but using only bounding boxes (from off-the-shelf algorithm, e.g., YOLOv7). 
As seen in Fig.~\ref{fig:real:robustness}(b), even such coarse approximations suffice for GeoPF to generate safe, efficient and real-time motions in clutter, while state-of-the-art methods Fig.~\ref{fig:real:robustness}(c)-(d) failed despite access to full geometry. }




Second, in a 3D task, the robot 
navigated around a shelf decomposed into multiple planar elements, along with additional obstacles. GeoPF successfully avoided all obstacles by generating environment-aware trajectories.

In the 
dynamic scene shown in Fig.~\ref{fig:teaser}, 
%
 the robot performed tasks like pouring tea into a moving cup or navigating around moving obstacles, with GeoPF consistently maintaining safe distances in real time without parameter adjustment. Across all cases, GeoPF produced real-time smooth, collision-free trajectories suitable for control-based execution.

\begin{figure}[t]
\centering
\begin{subfigure}{0.32\linewidth}
		\centering
		\includegraphics[width=1\linewidth]{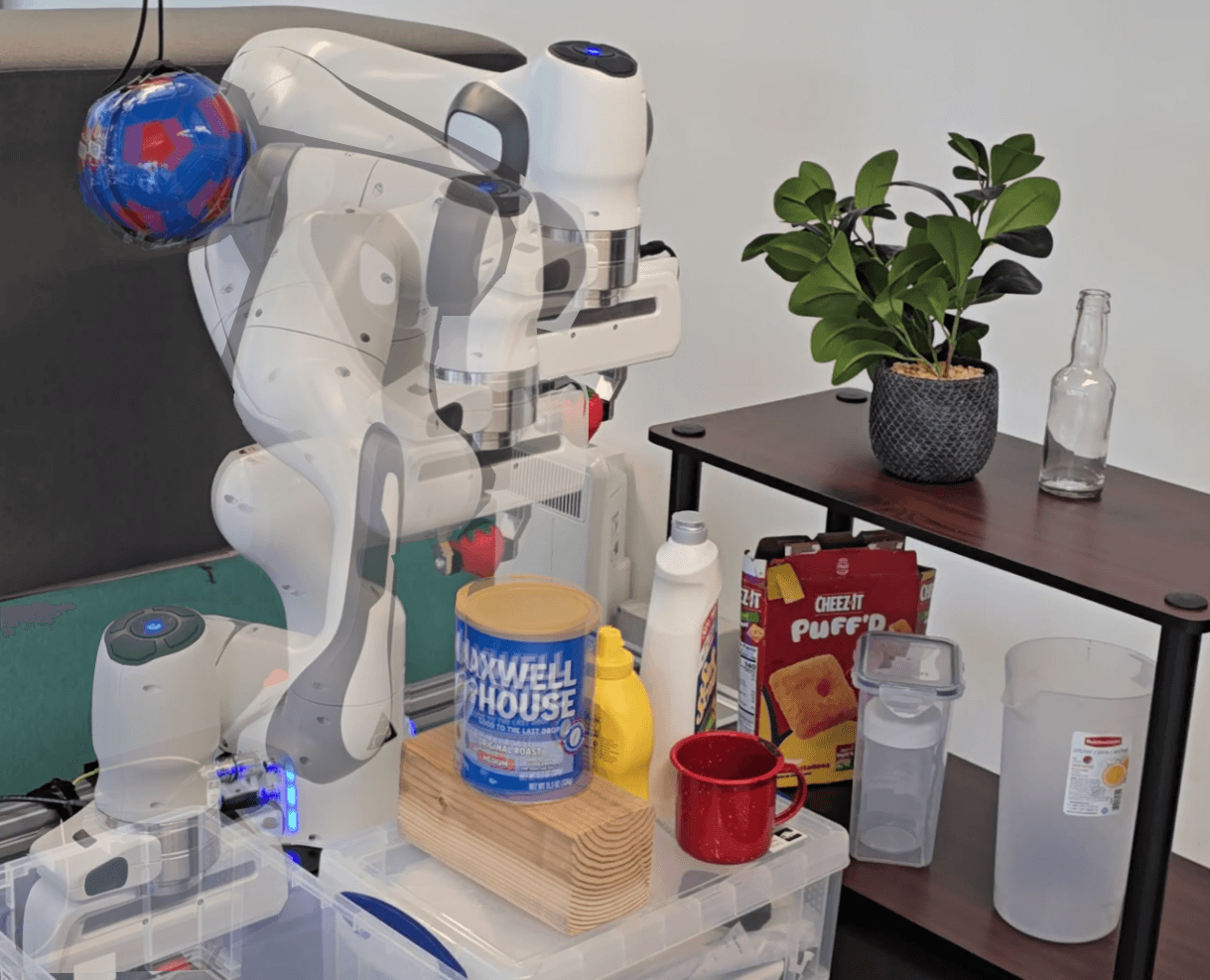}\vspace{-5pt}
		\caption{\scriptsize{Reactive (no avoidance)}}
            \label{fig:noavoi2}
	\end{subfigure}
\begin{subfigure}{0.32\linewidth}
		\centering
		\includegraphics[width=\linewidth]{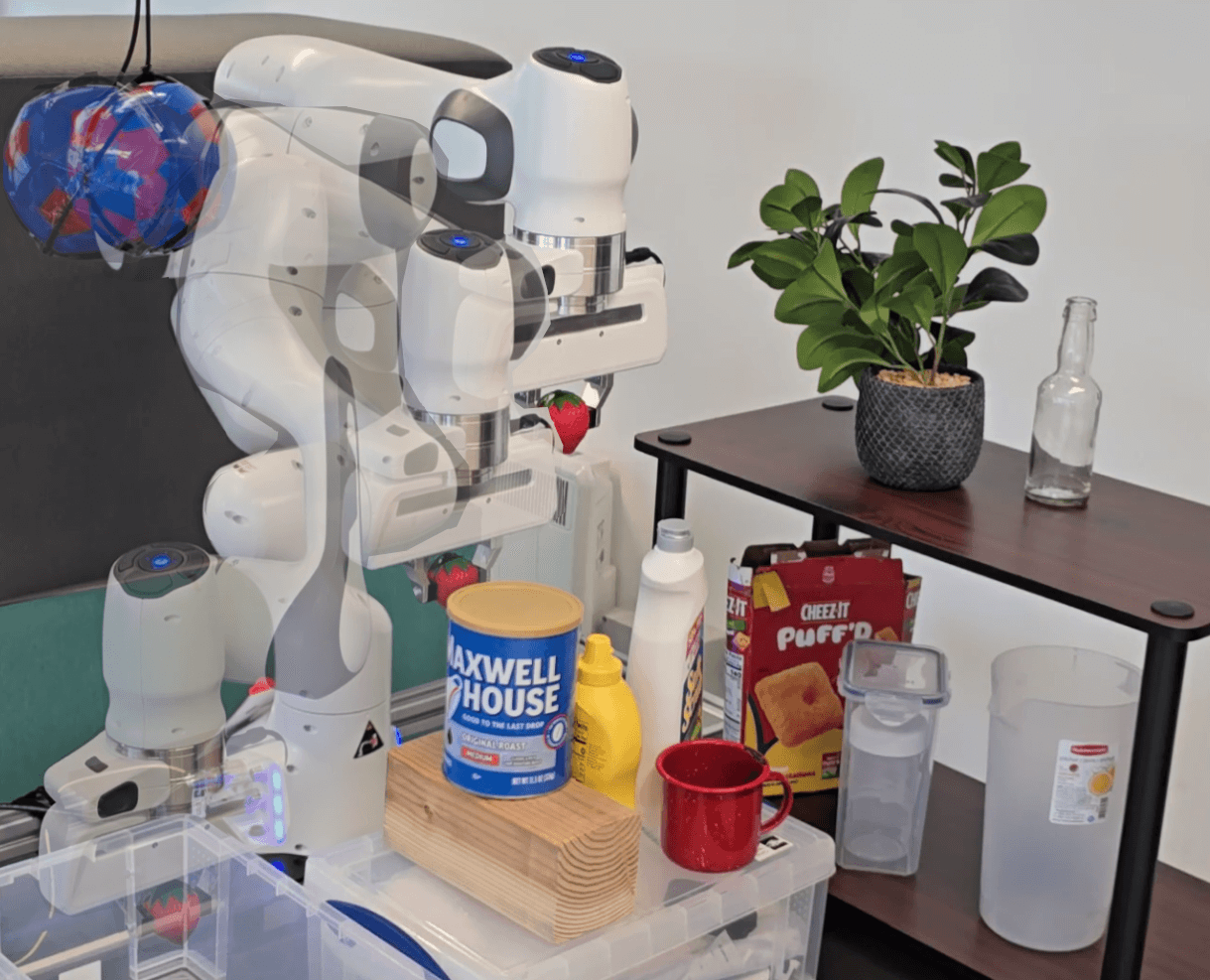}\vspace{-5pt}
		\caption{\scriptsize{GeoPF (w/o whole-body)}}
            \label{fig:ee2}
	\end{subfigure}
\begin{subfigure}{0.32\linewidth}
		\centering
		\includegraphics[width=\linewidth]{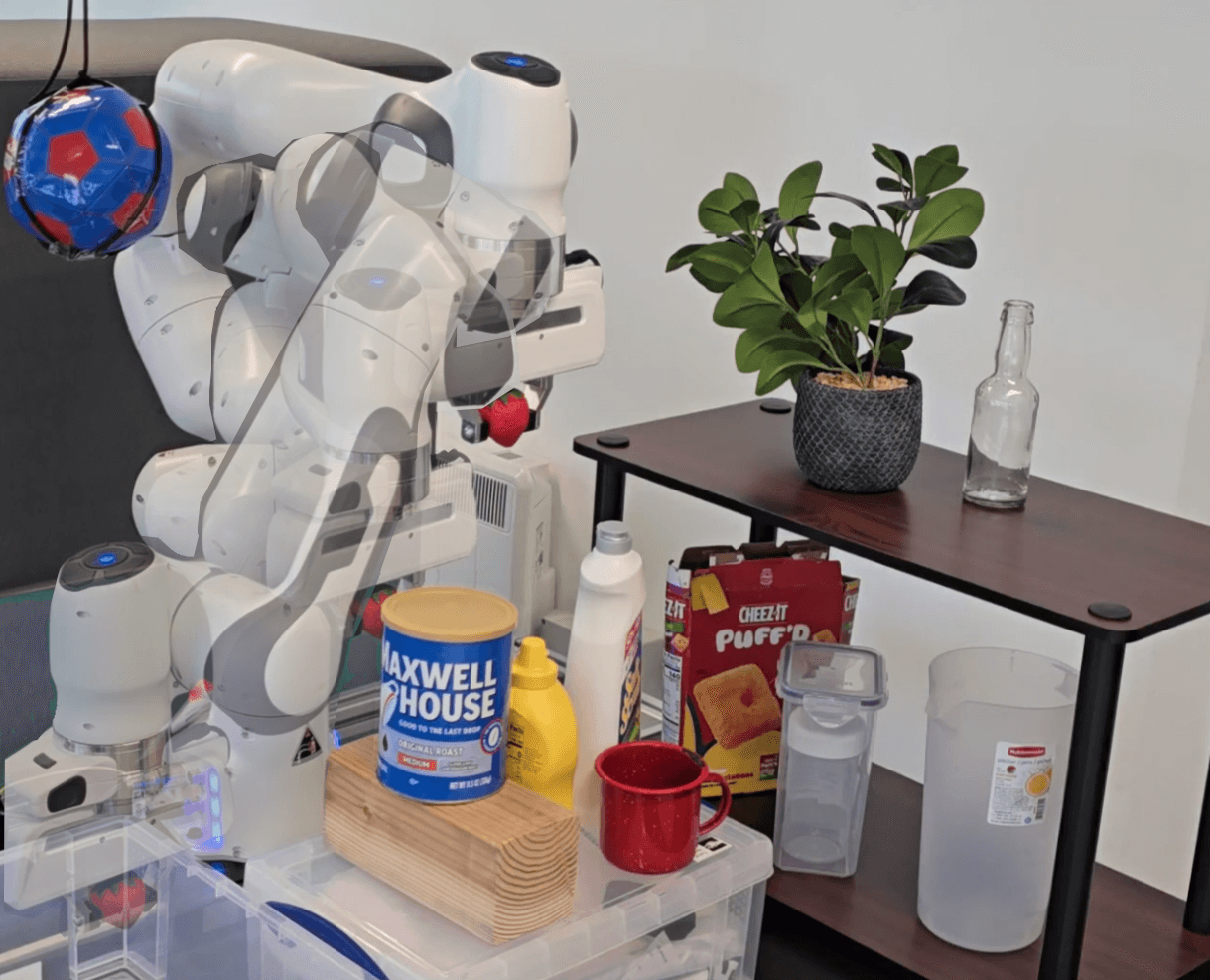}\vspace{-5pt}
		\caption{\scriptsize{GeoPF w/ whole-body)}}
            \label{fig:fullbody2}
	\end{subfigure}
\vspace{-2pt}
\caption{%
\revised{Whole-body collision avoidance with GeoPF. (a) Planner with no avoidance, naturally leading to multiple collisions, (b) End-effector-only avoidance fails to avoid elbow collisions. (c) GeoPF's whole-body geometric-aware collision avoidance successfully modulates arm and end-effector trajectories to safely avoid all obstacles.}
}\vspace{-5pt}
\label{fig::wholebody:avoidance}
\end{figure}

 \revised{
Lastly, we tested GeoPF for simultaneous EE and whole-body collision avoidance (Fig.~\ref{fig::wholebody:avoidance}). The links were shaped as cylinders, actively modulating reactive forces as detailed in Sec.~\ref{sec:compute:cylinder}. The resulting forces were then mapped to torques based on the moment arm to the previous joint. This enabled coordinated, geometry-aware avoidance  across the entire kinematic chain without additional tuning.     
 }


\section{Conclusion} \label{sec:conclusion}


This work introduces GeoPF, a novel reactive planner that unifies spatial structured geometric primitives under a temporal potential-field framework for real-time planning.  By explicitly leveraging closed-form distance functions \revised{and structural and spatial properties of primitives,} 
GeoPF delivers substantial improvements 
in computational efficiency, tuning simplicity, and collision avoidance performance.
Experiments demonstrate its robustness across static and dynamic scenarios, including whole-body avoidance, with minimal parameter adjustment.
Quantitative analysis and experiments demonstrate \revised{its reliability, robustness, and easy-deployment over existing field-based planners}, being the only suitable for real-time ($1$ kHz) performance in cluttered, unstructured dynamic settings. \revised{GeoPF has also been tested for whole-body avoidance and limited primitive-fitting access, all with no tuning.} 

%
%

Future work will focus on integrating prediction modules for dynamic obstacle tracking and multi-agent vector fields to address a core limitation of GeoPF—and all reactive planners—its lack of completeness. We also aim to fuse GeoPF’s primitive structure with neural SDFs (e.g.,~\cite{staroverov2024dynamic}), extending ANN inference to produce rich geometric descriptors.




%
%
%
\bibliographystyle{IEEEtran}
\bibliography{bibliography}

\end{document}